\documentclass[sigconf,preprint]{acmart}

\AtBeginDocument{%
  \providecommand\BibTeX{{%
    \normalfont B\kern-0.5em{\scshape i\kern-0.25em b}\kern-0.8em\TeX}}}

\def\semicolon{;}
\def\applytolist#1{
    \expandafter\def\csname multi#1\endcsname##1{
        \def\multiack{##1}\ifx\multiack\semicolon
            \def\next{\relax}
        \else
            \csname #1\endcsname{##1}
            \def\next{\csname multi#1\endcsname}
        \fi
        \next}
    \csname multi#1\endcsname}
\def\calc#1{\expandafter\def\csname c#1\endcsname{{\mathcal #1}}}
\applytolist{calc}QWERTYUIOPLKJHGFDSAZXCVBNM;
\def\bbc#1{\expandafter\def\csname bb#1\endcsname{{\mathbb #1}}}
\applytolist{bbc}QWERTYUIOPLKJHGFDSAZXCVBNM;

\usepackage{comment}

\usepackage{breqn}

\begin{document}

\title{Measuring, Interpreting, and Improving Fairness of Algorithms using Causal Inference and Randomized Experiments}

\author{James Enouen}
\email{enouen@usc.edu}
\affiliation{%
  \institution{University of Southern California}
  \city{Los Angeles}
  \country{USA}
}

\author{Tianshu Sun}
\email{tianshus@marshall.usc.edu}
\affiliation{ 
  \institution{{University of Southern California}}
  \city{Los Angeles}
  \country{USA} 
}

\author{Yan Liu}
\email{yanliu.cs@usc.edu}
\affiliation{%
  \institution{University of Southern California}
  \city{Los Angeles}
  \country{USA}
}


\begin{abstract}
Algorithm fairness has become a central problem for the broad adoption of artificial intelligence.
Although the past decade has witnessed an explosion of excellent work studying algorithm biases, achieving fairness in real-world AI production systems has remained a challenging task. 
Most existing works fail to excel in practical applications since either they have conflicting measurement techniques and/ or heavy assumptions, or require code-access of the production models, whereas real systems demand an easy-to-implement measurement framework and a systematic way to correct the detected sources of bias.

In this paper, we leverage recent advances in causal inference and interpretable machine learning to present an algorithm-agnostic framework (MIIF) to Measure, Interpret, and Improve the Fairness of an algorithmic decision.
We measure the algorithm bias using randomized experiments, which enables the \textit{simultaneous} measurement of disparate treatment, disparate impact, and economic value.
Furthermore, using modern interpretability techniques, we develop an explainable machine learning model which accurately interprets and distills the beliefs of a blackbox algorithm.
Altogether, these techniques create a simple and powerful toolset for studying algorithm fairness,
especially for understanding the cost of fairness in practical applications like e-commerce and targeted advertising, where industry A/B testing is already abundant.
\end{abstract}





\maketitle

\section{Introduction}

State-of-the-art AI systems are exceptional at modeling complex dependencies in data and are beginning to play a major role in devising solutions to many important tasks throughout society \cite{tai2015improved,alipanahi2015predicting,he2016deep,huang2017densely, vaswani2017attention}. While these models achieve significant improvement in prediction accuracy,
both the interpretability and the fairness of AI models have become critical obstacles to their widespread adoption \cite{barocas2016big,hardt2017fairness,kleinberg2018algorithmic,weller2019transparencyCall2Act}.

Recently, a body of excellent work has attempted to address the fairness issues of AI systems \cite{hardt16unidentifiable,joseph2016fairness,mitchell2018prediction,yao2017beyond,informs20faiSurvey}.  Some works focus on proposing general definitions of AI fairness, such as individual fairness \cite{kusner17counterfactualFairness} and group fairness \cite{dwork2012fairness}, while others target specific applications, including judicial decisions \cite{kleinberg2018human,cowgill2018impact}, job hiring \cite{cowgill2017automating}, advertising \cite{lambrecht2019algorithmic}, and language processing of online news \cite{bolukbasi2016man}. However, there are fundamental challenges in applying current research on algorithm fairness to real systems, including ambiguous definitions \cite{narayanan2018translation}, internal incompatibility \cite{kleinberg16tradeoffs}, difficulty in the interpretation of AI fairness \cite{doshi2017towards}, and lack of transparency for policy making \cite{kleinberg2018discrimination}. 
Specifically, for large-scale AI systems: (1) access to full code and training data is often impossible for external audiences and the auditing process; (2) there is a lack of implementable and scalable methods which can not only measure but also improve algorithm fairness, disentangling algorithm bias from dataset bias during improvement; (3) there is a lack of systematic ways to measure disparate treatments and disparate outcomes simultaneously, making it difficult to evaluate or balance fairness tradeoffs across separate stages. 
As a result, progress on fairness in industry AI systems is difficult to achieve and audit since no frameworks or methods are available for continuous steps of improvement beyond one-shot investigation of observational data.

In this paper, we leverage recent advances in economics, experimental design, causal inference, and explainable machine learning to propose a simple yet effective framework ``MIIF" for \textit{measuring, interpreting, and improving fairness} via randomized experiments to address these existing challenges. 
Our method effectively integrates randomized experiments and an interpretable model to be able to: (1) simultaneously measure treatment fairness, outcome fairness, and economic benefit for any given algorithm; (2) measure causal fairness without structural assumptions that are difficult to verify; (3) isolate algorithmic biases from dataset biases; (4) interpret a blackbox algorithm without access to code or training data, hence enabling fairness improvement in real-world systems.

\textcolor{black}{
By investigating how an algorithm's treatment decision is contingent on users' sensitive attributes, we can understand whether and how a machine learning algorithm is potentially creating biases and the way to eliminate such biases step-by-step.
Specifically, we apply a feature interaction detection model on top of randomized experiment data to identify the source of algorithm bias.
We leverage a generalized additive model (GAM) and a novel feature interaction detection method (Archipelago) to improve algorithm fairness by removing biased interaction terms in the algorithm.
}
\textcolor{black}{
In summary, by leveraging an interpretable model alongside randomized experiments, our new framework is set up to address many of the practical concerns of measuring, interpreting, and improving fairness in real production systems.
Continuous improvement can be achieved through the cycle of bias-identification and model comparison while being credibly communicated to an external audience, as is necessary for the iterative process of improving fairness in realistic pipelines.
}

In this work, we combine the randomized experiment framework with interpretability methods to be able to measure and improve fairness, yielding the following contributions:
\begin{itemize}
  \item We propose a unique approach to measure algorithm fairness using randomized experiments, which enables the simultaneous measurement of disparate treatment, disparate impact, and economic value.
  \item We provide a novel framework incorporating \textit{feature interaction detection} and \textit{generalized additive models} to build an interpretable and flexible model type which accurately distills blackbox models in a data-efficient manner.
  \item We develop a novel modification to any existing group fairness metrics which converts them into causal-aware fairness metrics. 
  We call the metric ``no-worse-off'' and benchmark against an interpretable, audited algorithm.
  \item We investigate a setting which moves away from observational and non-causal fairness metrics without requiring heavy causality assumptions from the researcher, allowing for new insights on the tradeoffs between economic value and fairness performance. 
\end{itemize}

\begin{figure*}[t]
\centering
\includegraphics[width=0.33\textwidth]{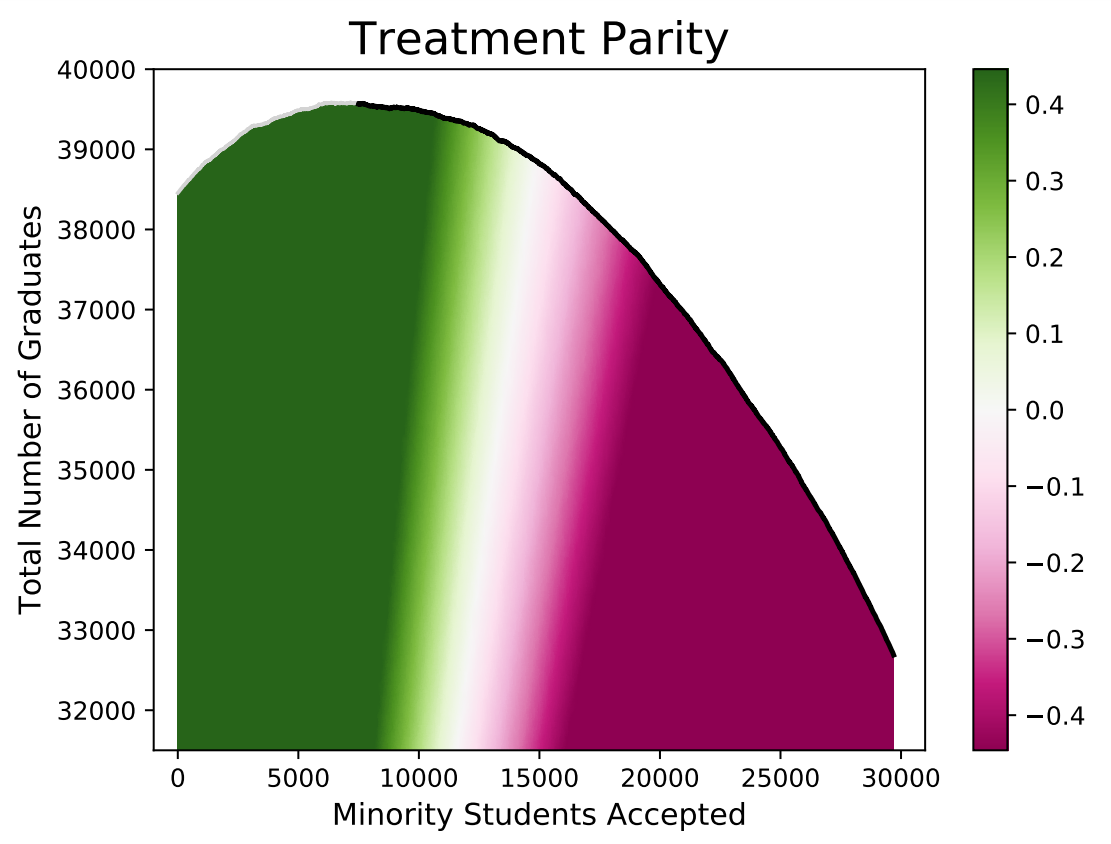} 
\includegraphics[width=0.333\textwidth]{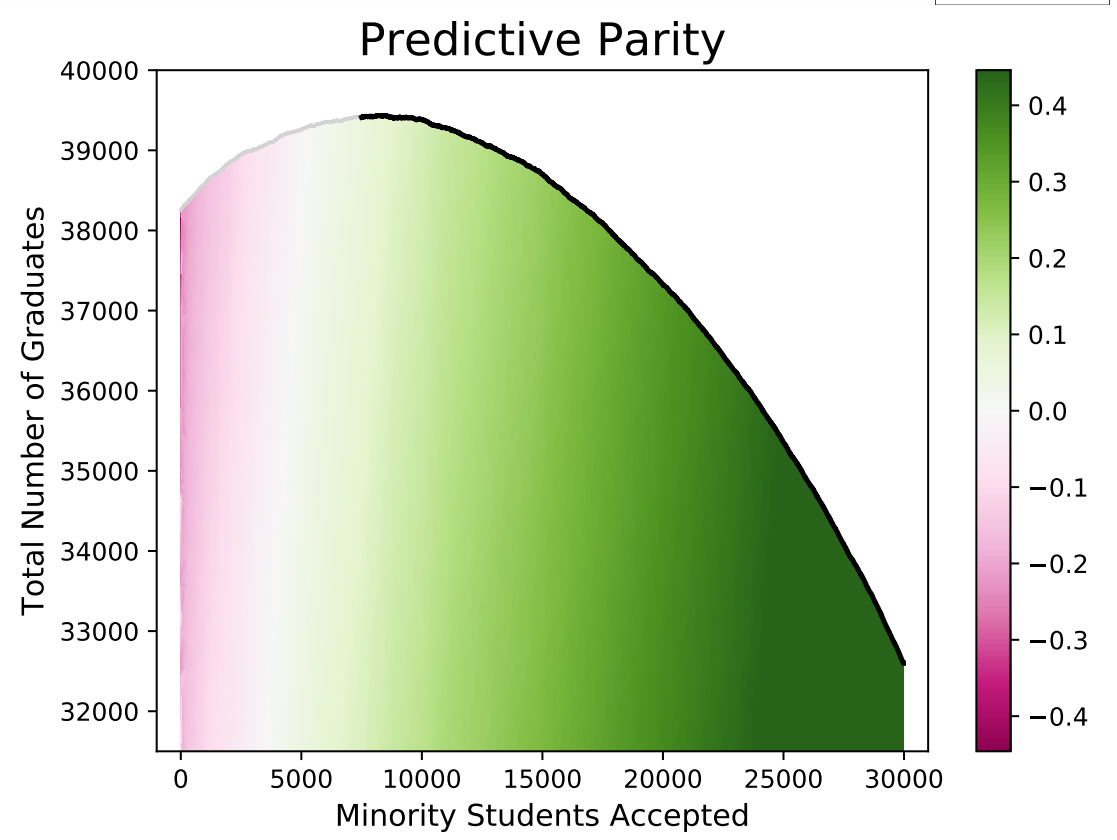}
\includegraphics[width=0.324\textwidth]{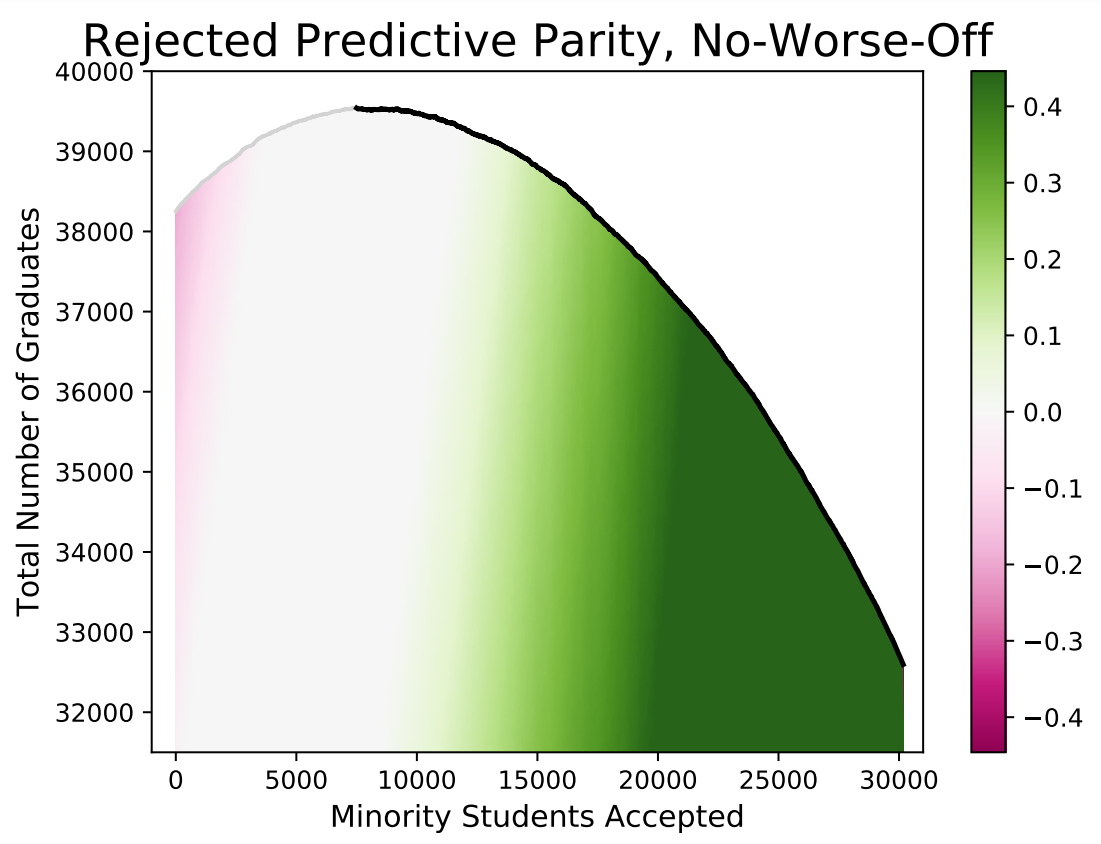}
\caption{ College Admission Dataset. Treatment Parity, Predictive Parity, and No-Worse-Off for Rejected Applicants. \\
\textnormal{ Three plots of the region of feasible college admission policies plotted on the two axes of `total number of graduates' and `total number of minority students admitted'.  For each plot, the region is shaded according to value taken by its corresponding fairness metric.  Values of zero correspond to perfect fairness and the sign represents the direction of unfairness.  The optimal Pareto frontier along the top-right boundary of the region is shaded in black.}
}
\label{fig:college_different_metrics}
\end{figure*}

\section{Background and Related Work}
In this section, we provide background on the three key domains we consider in this work: algorithm fairness, causal inference, and interpretability.
We first give a fairly extensive overview of the main topic of algorithm fairness.
Then, using an illustrative example from the literature, we highlight some of the major shortcomings of existing fairness work.
Finally, we introduce the relevant topics from both causal inference and interpretable machine learning which allow us to address many of these shortcomings.

\subsection{Algorithm Bias and Fairness}
Modern study of the fairness of computational algorithms was {rekindled} when ProPublica \cite{propub16} uncovered biases in the COMPAS algorithm for crime prediction. 
Similar discrepancies for \textit{sensitive features} like race, gender, and age have been discovered in algorithms across many applications including loan approval, job screening, targeted advertising, and facial recognition. 
In recent years, there has been significant development on the study of algorithm fairness.
However, appropriately addressing fairness of AI models remains a significantly challenging problem, especially for industry-scale systems.

\subsubsection{Observational Metrics}
There are many measures of fairness which can be readily applied to existing datasets:
treatment parity, predictive parity, equal opportunity, and equalized odds, just to name a few.
Unfortunately, these observed metrics are also accompanied by an abundance of conflicts in their compatibility; it is rare that two observational metrics are simultaneously achievable \cite{kleinberg16tradeoffs,dwork11individualFairness,berk17sixConflicting,liu2019implicitCalibration}.
Further, it is now well-known that two completely different causal models can lead to indistinguishable distributions from an observational perspective, leading many to firmly believe fairness is not achievable without causal reasoning \cite{hardt16unidentifiable,kilbertus17causalfairness}.

\subsubsection{Causal Metrics}
The abundance of negative results in observational metrics have pushed researchers towards approaches which make rigorous assumptions on the causal structure (often in the form of a structural causal model or SCM) in order to detect bias \cite{neto20causalfairness,Chiappa_2019_pathSpecCtfFair,Zhang_Bareinboim_2018_fairness_decision_making,salimi2019interventionalFairness,kusner17counterfactualFairness,makhlouf2022causalFairSurvey}.
Amongst the most popular of these methods are counterfactual fairness and path-specific counterfactual fairness \cite{kusner17counterfactualFairness,zhang2017directIndirectPathsFairness,Nabi_Shpitser_2018_fairInference,wu2019pathSpecCtfFair}. 
Unfortunately, these definitions often shift the burden to the data scientist: making the difficult and socionormative decisions to build a full causal graph between all of the sensitive and nonsensitive variables.
This ranges from challenging on a handful of features to completely unfeasible on industry-scale problems.
Further, misspecifying the SCM can easily result in incorrect fairness conclusions \cite{kilbertusBall2019sensitivityCtflFairConfounding,philip_Ball2018thesis}.

More recent work has only exposed further shortcomings of these causal notions of fairness.
Counterfactual fairness has been shown to be an unidentifiable and inadmissible quantity while also failing to capture background effects
\cite{pleckoBareinboim2022causalFairnessAnal}.
Further, it is often deemed unreasonable to intervene directly on a sensitive attribute, which is intimately tied to its societal context and related consequences \cite{huLily2020whatsSexGotToDo,khademi2019fairness_decision_making}.
The work in \cite{nilforoshan2022causalConceptionsFairness} has shown that even with access to the true SCM, counterfactual fairness can `overcorrect' the world to be completely fair, forcing suboptimal decisions, even for those stakeholders who are actively interested in increasing diversity.

\subsection{Motivating Example}
In order to make these notions clearer, we reintroduce the simple model for college admissions focused on in \cite{nilforoshan2022causalConceptionsFairness}.
Here, we assume that there is a majority and minority group  whose test scores are correlated with group membership.
The college admission board is then tasked with balancing the two competing objectives of maximizing four-year graduates and maximizing diversity.
For further details on this example, see Appendix \ref{app_sec:datasets} or \cite{nilforoshan2022causalConceptionsFairness}.

In Figure \ref{fig:college_different_metrics} below, we can see an array of different policies depicted with their final outcomes for the two objectives of interest: four-year graduates and minority acceptances.
For all plots, we can see the black curve in the top-right corresponding to the Pareto frontier which optimally balances these two objectives.
Each plot is additionally shaded by its corresponding fairness metric, with zero representing no unfairness and the sign representing the direction of unfairness.

Starting with treatment parity, we can see that there indeed exists an optimal policy which achieves perfect treatment parity.
However, this metric alone provides no further flexibility to achieve the other Pareto optimal policies and provides no guidance on how to achieve other optimal policies.
The story is even worse for the metric of predictive parity, which is achieved by none of the Pareto optimal policies.
Comparing between these first two plots of Figure \ref{fig:college_different_metrics}, one can also see the general incompatibility for treatment parity and predictive parity to be satisfied simultaneously.

It is further known from \cite{nilforoshan2022causalConceptionsFairness} that none of the five causal metrics they study achieve Pareto optimal policies either, namely: counterfactual fairness, path-specific fairness, counterfactual predictive parity, principal fairness, and counterfactual equalized odds.

Altogether, these theoretical and empirical results highlight the need to be able to make fair decisions, even in an unfair world.
In this work, we instead focus on a known but underexplored perspective that algorithm unfairness can only be created out of the algorithm's decisions themselves.
Furthermore, the causal impact these decisions have on the lives of their constituents is the only reasonable notion for measuring unfairness.

Later, in Section \ref{sec:fairness_definitions}, we define the metric of NWO ("no-worse-off") when compared to a baseline treatment.
Looking at our college admission example, we compare (a) the rate of graduates per group if the college were not to exist; against (b) the rate of graduates per group under the college's potential decision algorithm.
Here, we argue that if a college were to accept no students, this would only reflect the natural --but perhaps unfair --state of the world.
Accordingly, comparing any fairness metric against the value achieved in such a counterfactual, baseline world allows for additional flexibility which is not plausible in an purely observational setting.
Indeed, in Figure \ref{fig:college_different_metrics}, we can see that achieving perfect NWO is compatible with an entire collection of different Pareto optimal policies.

\subsection{Causal Inference}
In order to reinforce our perspective that unfairness is caused by the treatment of the algorithm, we draw from the existing literature on causal inference.
In this work, we focus on the Neyman-Rubin causal model \cite{rubin1983causalmodel,holland1986causalmodel} which is the model used for medical trials, policy interventions, and economic planning.
Causal inference is designed to estimate how much a treatment (such as an algorithmic decision) affects an individual's final outcome (such as college graduation, recidivism, or blood donation).
We are particularly interested in how the effects of a treatment on an individual's outcome depends on their personal covariates and sensitive features.

Causal inference has long been a mainstay of policy choice and algorithm fairness should likely be no different.
Previous sociological discussions \cite{barabas17interventions} have long argued that machine learning in criminal justice should instead shift to this viewpoint considering the causal effect of a change in policy or algorithm,
because this perspective focuses on risk mitigation instead of risk prediction.

{Randomized experiments} or clinical trials refer to the scenario where each member of the initial population was assigned a truly random treatment, allowing us to unambiguously identify the effect of the treatment.
This removes the concerns of hidden dependencies and confounding variables, which can bias the model estimation procedure.
Despite their myriad benefits, their primary drawback is their costly set-up in a real-world environment.
However, random experiments have become rather standard in many existing commercial AI systems, such as advertisement ranking systems, recommender systems, and other algorithmic pipelines.

Consequently, we see this intersection between fairness and causal inference as a {potential breeding ground} for future ideas in algorithmic fairness where causal-based experiments can be run for relatively cheap.
We further note that even without access to large-scale randomized experiments, our framework can still be adjusted to yield accurate estimates under multiple realistic settings.
In high-stakes regimes like personalized medicine and recidivism prediction, one is only able to perform algorithmically random experiments: randomly switching between two previously vetted decision algorithms.
Propensity scoring and reweighting can easily be leveraged in this context since we can be sure there are no hidden confounders in our setup \cite{rubin1983causalmodel}.

\begin{figure*}[h]
    \centering
    \includegraphics[width=1.39\columnwidth]{    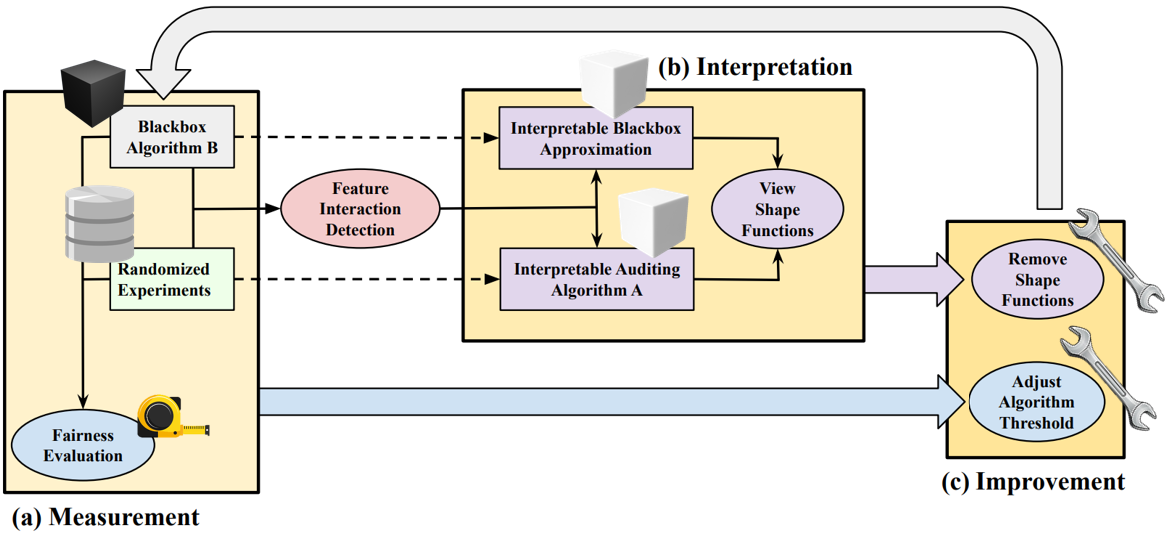}
    \caption{Diagram of the entire MIIF framework.
    \\ \textnormal{\small (a) the first phase `measurement' of the ITE model using random experiments for fairness evaluation (b) the second phase `interpretation' of the underlying biases through the interpretable GAM model (c) the third phase `improvement' of fairness using both GAM shape techniques and thresholding, updating the  blackbox model and reevaluating with random experiments, continuing the cycle of improved fairness.}
    }
    \label{fig:full_diagram}
\end{figure*}

\subsection{Interpretable Machine Learning}
Yielding accurate and consistent interpretations of machine learning models has many important consequences for trustworthiness, auditing, and robustness.
In fact, many of the goals of interpretable machine learning are aligned with those of algorithm fairness: both demand that models or algorithms make reasonable decisions and do not succumb to arbitrary biases.
In this work, we will leverage interpretability as a tool to ensure fairness.

Many popular explanations methods like LIME focus on local explanations of a blackbox model; however, these methods can be unfaithful to the global structure
\cite{ribeiro2017lime,dhamdhere2019shapley,sundararajan2017axiomatic,janizek2020explaining}.
In fact, it has been shown that blackbox models can be adversarially trained to disguise unfairness from these local explanation methods 
\cite{anders2020fairwashingOffManifold,dimanov2020shouldntTrustExplanations}.

Our work instead focuses on global explanations using a combination of additive models \cite{hastie1990originalGAM,lou2013accurate} and feature interactions \cite{tsang2017detecting,sorokina2008detecting}.
We combine an interpretable model called a generalized additive model with modern feature interaction detection techniques \cite{tsang2020archipelago} to stay interpretable without losing predictive power.
We then use this interpretable model to explain why blackbox models are making their decisions, extending distillation approaches like \cite{tan2018learning} to respect feature interactions while also being more data efficient.


\section{Methods}
In this section, we introduce our simple but effective framework, namely Measurement, Interpretation, and Improvement of Fairness (MIIF) which leverages the recent developments in explainable machine learning, economics, and causal inference. 
It consists of three components (1) measuring  unfairness in AI systems using randomized experiment to characterize the differences in the treatment and outcome at an individual level, (2) interpreting the complex neural network models via a state-of-the-art interaction detection method to construct a generalized additive model which can be fully interpreted, and (3) improving the prediction models based on the identified source of bias with generalized additive models and mock experiments.
The framework is depicted below in Figure \ref{fig:full_diagram}.


\subsection{Measurement: Two Independent Models}


In our notation,
each individual is described by their $d$-dimensional features 
$X\in\cX\subseteq\bbR^d$, their received treatment $T\in\{0,1\}$, and their final outcome $Y\in\bbR$.
The fundamental problem of causal inference is that we can only ever observe one of the two `potential outcomes': the outcome when receiving the treatment ($Y_1$) or the outcome without receiving the treatment ($Y_0$), depending on the treatment received by the individual ($T$).
As is often convention, we will implicitly denote $Y = Y_T$.
Further, in our study of fairness, we consider that one dimension of $X$ is a sensitive binary feature\\ $S\in\{0,1\}$.
We also consider a decision algorithm $D:\cX\to\{0,1\}$, which makes a treatment decision based on an individual's covariates.
Although our framework can easily be extended beyond binary sensitive features and binary treatments, we limit our discussion to these cases for clarity.

Our goal in causal inference is then to estimate how the outcome $Y$ depends on the covariates $X$ and treatment $T$.
The individual treatment effect (ITE) or conditional average treatment effect is defined as the difference in expected value by choosing treatment T=1 over treatment T=0 as a function of the observed covariates:
 \begin{equation} 
ITE(x) = \mathop{\mathbb{E}[Y|X=x,T=1] - \mathbb{E}[Y|X=x,T=0]} \label{eq:ITE}
\end{equation}
\noindent
Historically, researchers only conditioned on a handful of features, or no features at all (where this quantity is called the average treatment effect.)
With the modern power of machine learning, however, it is possible to condition on increasingly high-dimensional covariates
as in targeted advertising and personalized medicine.

As was previously mentioned, randomized experiments is a powerful framework that allows us to forgive the interplay between the treatment variable and covariate variables by fitting two separate models in the ``T learner'' setup, following the works of \cite{McFowland2020optimal, athey2017econometrics,wager2018estimation}. 
The machine learning models we will consider in our results are deep neural networks, random forests, boosting methods, support vector machines, and generalized additive models.

\subsection{Interpretation: Feature Interactions}
We introduce the generalized additive model we use as our global interpretable model and briefly discuss how we use this to distill the trends learned by the blackbox model.

\paragraph{Generalized Additive Models}
We consider the generalized additive model (GAM), a generalization of linear regression \cite{hastie1990originalGAM,agarwal2020nam}.
\begin{equation}\label{eq:GAM2}
y = \sum_i f_i(x_i) + \sum_{i<j} f_{i,j}(x_i,x_j)
\end{equation}
Using only the first term refers to the classical, univariate GAM we denote GAM1.
The second term extends GAMs to bivariate models which can represent feature interactions, and which we will call GAM2.
We note that including all feature interactions into the additive greatly increases the modeling capacity, allowing us to more accurately capture the blackbox predictions.
However, alongside this benefit of increased capacity is the consequence of increased data requirements, which is potentially unattainable for most realistic auditing pipelines.

Accordingly, we leverage a novel interaction detection procedure \cite{tsang2020archipelago} to rank all possible feature pairs, including only those which are ranked highest.
This allows us to drastically cut down on our distillation data requirements while also obtaining highly flexible modeling capacity.
Further details are described in Appendix \ref{app_sec:feature_interactions}. 

We then fit the interpretable model to (a) accurately fit the auditing dataset, and (b) accurately reproduce the blackbox model's predictions, distilling its predictions into an interpretable form.
In our results section, we consistently find that our distilled interpretable model is an accurate approximation of the true blackbox neural network predictions.
By contrasting the learned shape functions between the distilled model and the audit model, we can understand the reasoning behind the blackbox algorithm and how it differs from the trends learned from the data biases alone.

\subsection{Improvement: Removing Sensitivities }
In our experiments, we focus on improving the fairness of a blackbox model using the following two simple methods.
The first is the extremely popular and general method of multiple thresholds.
The second is a technique designed specifically for our interpretability method.
We leave further investigation of the effects of other improvement methods to future work.

\subsubsection{Multiple Thresholds}
We use the well-known method of assigning multiple thresholds to each protected group in order to alleviate the differences in treatment over two or more populations.
Interestingly, because our work has access to random experiment data, each threshold corresponds to a decision algorithm for which we can run a mock experiment based on each individual's assigned treatment.
Evaluating on the subset of individual's who received their real-world treatment is unbiased because of the randomization of the experiment.
Altogether, this generates a manifold of different decision outcomes before deployment and is {what gives our work a unique perspective} on the price of fairness, as seen in Figure \ref{fig:3d_fairness_1_model}.

\subsubsection{Sensitive Feature Functions}
After we have learned the simple, low-dimensional trends of the blackbox model, we can remove the trends which depend on a sensitive feature.
In doing so our model first learns the dependence on a sensitive feature and then explicitly removes the learned effect.
Unlike fairness through awareness, our algorithm learns the underlying trend and then actively chooses to remove it, rather than potentially focusing on correlated proxies.
As one might expect, we generally find that this improves the treatment fairness of the model since it becomes blind to the impact of having such a sensitive feature.
In addition to completely removing the trend, we also replace the shape function with the one which is learned by the interpretable model trained on an auditing dataset.
We find this also improves the fairness of the decision algorithm.

\subsection{Evaluate: Measurement of Fairness}
\label{sec:fairness_definitions}
In this paper, we will use treatment fairness and outcome fairness as the two primary measurements for fairness.
\paragraph{Treatment Fairness}
For treatment fairness, we simply ask for `statistical parity' on the treatment variables.
That is to say we ask that the same percentage of each group is given the treatment.

\begin{equation} \label{eq:treat_fair}
\bbP(T=1 | S=1) = \bbP(T=1 | S=0)
\end{equation}

\paragraph{Outcome Fairness}
For outcome fairness, we use a measure similar to `predictive parity' where, amongst the treated members, we have the same expected outcome across each of the groups.

\begin{equation} \label{eq:outcome_fair}
\bbE[ Y_1 | T=1, S=1] = \bbE[ Y_1 | T=1,S=0]
\end{equation}

\paragraph{Fairness Metrics}
We can now evaluate a decision algorithm $D:\cX\to\{0,1\}$, which decides whether or not to treat each individual, using the following metrics:
\begin{equation} \label{eq:TF}
TF(D) := \frac{\bbE_{X^{(0)}}[D(X^{(0)}) | S=0]}{\bbE_{X^{(1)}}[D(X^{(1)}) | S=1]}, \;\;\;\;\;
\end{equation}  
\begin{equation} \label{eq:OF}
OF(D) := \frac{\bbE_{X^{(0)}}[Y_1 | S=0,{D(X^{(0)})}=1]}{\bbE_{X^{(1)}}[Y_1 | S=1, {D(X^{(1)})}=1]}
\end{equation}

These metrics result from taking the ratios of the previous equations (\ref{eq:treat_fair}) and (\ref{eq:outcome_fair}) and plugging in the decision algorithm $D$ for the treatment $T$.
We evaluate both of these parity equations using the $p\%$-rule for binary features.
This measurement simply takes the ratio between the two values for each sensitive class, inverting the ratio if necessary to remain $\leq 1$.
A score of $100\%$ then corresponds to perfectly satisfying the above equalities.
This metric comes from the current legal rule-of-thumb which claims anything below $80\%$ is possibly discrimination. 
We will now refer to these two metrics as $TF$ and $OF$ accordingly.



\paragraph{\textcolor{black}{Algorithm Comparison}}
We also introduce a contrastive metric specific to our framework we call "no worse off".
Here, we are able to compare an algorithm $D_B$ against a benchmarking algorithm $D_A$.
For example, we can consider ``How many students would still graduate college if we accepted no applicants this year?" or ``How many citizens would naturally donate blood if we did not remind any of them?"
In this way, we can compare against the baseline of no treatment, which could be argued to be guaranteed to be fair, since it represents the natural state of the world.
Alternatively, we can choose to benchmark against any other decision algorithm $D_A$ which we believe to be fair.
In this work, we focus on $D_A$ which are simple and interpretable to enable a comparison to a blackbox model.
We define No-Worse-Off for Outcome Fairness as follows:

\begin{equation} \label{eq:NWO_PP}
NWO(D_B,D_A) = \min{\bigg\{ \frac{OF(D_B)}{OF(D_A)}, \hspace{0.3em} 100\% \bigg\}}
\end{equation}

Importantly, this new definition depends critically on who we define as the population of interest and what algorithm $D_A$ we define as a reasonable comparison.  We delay a detailed discussion of these concerns to Section \ref{sec:ethical_NWO}.

Here, we instead choose to clarify a potential point of mild confusion, surrounding the two different types of treatments considered within this work.
The first type of treatment is the physical treatment of the corresponding dataset.
For our main blood donation dataset, this corresponds to being asked to donate blood.
Our second treatment type is the algorithmic treatment of using a decision algorithm $D_B$ instead of another algorithm $D_A$.
The latter setup is what we will use to contrastively compare blackbox algorithms against an audited counterpart, as in what is seen in the NWO definition.
The randomized experiment dataset to follow are of the former type and our causal fairness analyses are of the latter type.

\section{Experiment Datasets}
\label{sec:datasets}
We performed our experiments on four different datasets from a diverse set of fairness applications, highlighting the versatility of our setup.
We perform experiments on two synthetic datasets as well as two real-world datasets.
We provide a short summary of each dataset here, but for space we reserve a fully detailed explanation to Appendix \ref{app_sec:datasets}.


\textbf{College Admission Dataset}
We use an existing simulation of college admissions detailed in \cite{nilforoshan2022causalConceptionsFairness}.
The dataset focuses on a majority race with better access to test preparedness and a minority race with lesser access.
The college admission board must then make a decision based on race and test score alone.
The original work focuses on the shortcomings of five causal metrics for fairness.
\par

{\textbf{Synthetic Dataset.}}
We created this dataset to explicitly study how strong correlations between a sensitive attribute and an informative feature will affect fairness.
Four variables are sensitive features, with varying levels of correlation to eight additional variables which influence the final outcome.
The simulation obeys a simple set of structural equations detailed in Appendix \ref{app_sec:datasets}.
We focus on the impact that varying the correlation, $c$, has on fairness.

{\textbf{Blood Donation Dataset}}
This dataset was collected by a blood bank whose goal was to determine the efficacy of mobile messages in targeting individuals for donation.
The treatment corresponds to receiving a text message invitation to donate blood in exchange for a small prize.
The members of the baseline receive no such text message.
The outcome is whether the individual chooses to donate and how much blood the individual chooses to donate.
The covariates of the individual correspond to a number of sensitive features like  gender and age as well as nonsensitive features like blood type and donation history.
In total there are 34 covariates for 60,000 potential donors.

{\textbf{Collage Scrapbook Dataset}}
This dataset was collected by the online retailer who attempted to find the cost effectiveness of a referral program.
The treatment corresponds to sending an individual an email offering them a scrapbooking gift for every referral they make.
The outcomes are whether the individual refers a friend, whether they redeem their gift, and how much their friend spends on the new platform.
The covariates of each individual again correspond to sensitive features (gender, age) and nonsensitive features (purchase history).
There are 10 different features for 100,000 users.

\section{Results}
We first provide an evaluation of both the performance and the fairness of the different machine learning algorithms we consider on our real-world datasets.
Second, we interpret the results of our blackbox algorithms in comparison to our interpretable benchmarks.
Next, we look at a variety of techniques for adjusting and amending these blackbox algorithms, as well as understanding the benefits and drawbacks of each strategy for improving fairness.

\subsection{Measurement}
The first set of experiments we consider is our blood donation dataset.
The `donation percentage' refers to the percentage of people who donated in the program evaluation.
The `economic benefit' refers to the economic value earned by the program evaluation as described in detail in Appendix \ref{app_sec:datasets}.
The area under the curve (AUC) metric is also provided for both subsets of the dataset, treatment T=0 and treatment T=1.
We also measure TF and OF for both protected categories of gender and age.

\begin{table}[h!]
    \begin{centering}
    \caption{Blood Donation Measurement Results}
    \label{tab:blood_measurements}
    \small
    \scriptsize
    \resizebox{.48\textwidth}{!}{%
    \begin{tabular}{|c|c|c|c|c|c|c|c|c|} \hline
         & donation & economic & \multicolumn{2}{c|}{AUC} & \multicolumn{2}{c|}{gender} & \multicolumn{2}{c|}{age}  \\ 
         & percentage & benefit (RMB) & \multicolumn{1}{c}{T=0} & \multicolumn{1}{c|}{T=1} & \multicolumn{1}{c}{TF} & \multicolumn{1}{c|}{OF} & \multicolumn{1}{c}{TF} & \multicolumn{1}{c|}{OF}  \\
         \hline
None    & $0.69\%$ & 2.57$\pm$0.33 & -- &  -- &  -- &  -- &  -- &  -- \\ 
Random    & $0.92\%$ & 2.74$\pm$0.45  & -- &  -- &  $97.5$ &  $62.8$ & $98.3$ &  $57.5$ \\
\hline
SVM        & $1.03\%$ & 2.69$\pm$0.29  & $.541$ &  $.672$ &  $94.4$ &  $64.8$ &   $90.7$ &  $64.6$ \\ 
RF         & $1.12\%$ & 2.88$\pm$0.23  & $.591$ &  $.696$ &  $99.3$ &  $66.3$ &  $99.0$ &  $64.4$ \\ 
XGB        & $1.03\%$ & 2.90$\pm$0.37  & $.587$ &  $.716$ &  $85.7$ &  $74.0$ &  $95.4$ &  $64.8$ \\ 
DNN        & $1.04\%$ & 2.97$\pm$0.40  & $.597$ &  $.715$ &  $87.0$ &  $62.0$ & $84.5$ &  $62.7$ \\ 
\hline
GAM1       & $1.09\%$ & 3.07$\pm$0.57  & $.603$ &  $.661$ &  $90.9$ &  $66.7$ &   $90.3$ &  $61.8$ \\ 
GAM2    & $1.09\%$ & 3.09$\pm$0.38  & $.601$ &  $.676$ &  $91.3$ &  $64.5$ &  $84.8$ &  $58.0$ \\ 
\hline
    \end{tabular}} \\[10pt]
    \end{centering}
\end{table}

\vspace{-1em}
In the results from the blood donation dataset in Table \ref{tab:blood_measurements}, we can see how typical machine learning models such as random forests, boosting machines, and DNNs are able to accurately fit the dataset and yield an increase in economic performance over random treatment and AUC scores above 50\%.
We note that the GAM1 and GAM2 models are able to perform competitively alongside these blackbox algorithms, demonstrating their capacity to learn the trends of this dataset.
For both sensitive features of gender and age, we see that outcome fairness is consistently difficult to achieve.

In our Collage referral results below, we use `referral percentage' to describe the percentage of people who referred at least one friend in the program evaluation and `economic benefit' to describe to the economic valuation in dollars as described in the appendix. 

\begin{table}[h]
    \begin{centering}
    \caption{Collage Referral Measurement Results}
    \label{tab:collage_measurements}
    \small
    \scriptsize
    \begin{tabular}{|c|c|c|c|c|c|c|} 
        \hline
         & referral & economic &\multicolumn{2}{c|}{AUC} & \multicolumn{1}{c|}{gender} & \multicolumn{1}{c|}{age}  \\ 
         & percentage & benefit (\$) &\multicolumn{1}{c}{T=0} & \multicolumn{1}{c|}{T=1} & \multicolumn{1}{c|}{TF} & \multicolumn{1}{c|}{TF} \\
        \hline
None       & $3.31\%$ &  0.256$\pm$0.061 & -- &  -- &  -- &  --  \\ 
Random       & $3.47\%$ &  0.264$\pm$0.043 & -- &  -- &   $96.7$ & $94.9$   \\ 
\hline
SVM        & $3.54\%$ &  0.276$\pm$0.054 & $.696$ &  $.558$ & $94.1$ & $95.4$   \\ 
RF         & $3.49\%$ &  0.266$\pm$0.043 & $.660$ &  $.751$ & $98.3$ & $95.7$   \\ 
XGB        & $3.52\%$ &  0.277$\pm$0.035 & $.780$ &  $.742$ & $71.6$ & $85.3$   \\ 
DNN        & $3.45\%$ &  0.273$\pm$0.046 & $.763$ &  $.818$ & $66.1$ & $92.4$   \\ 
\hline
GAM1       & $3.46\%$ &  0.260$\pm$0.046 & $.758$ &  $.813$ & $90.4$ & $88.1$   \\ 
GAM2     & $3.50\%$ &  0.250$\pm$0.035 & $.757$ &  $.816$ & $91.4$ &  $86.4$   \\ 
\hline
    \end{tabular} \\[10pt]
    \end{centering}
\end{table}

\vspace{-.8em}
We again find that our models can accurately fit the data with high AUCs and yield positive economic benefit.
The SVM and XGB models perform marginally better than the other machine learning models we considered, but all are able to improve over the default lack of treatment.
We also note that while the GAM1 and GAM2 match or outperform other models in terms of AUC and referral percentage, they fall slightly behind in economic benefit.
Again, we observe varying levels of fairness achieved
by different algorithms.

\begin{figure*}[h]
\centering
\includegraphics[width=0.23\textwidth]{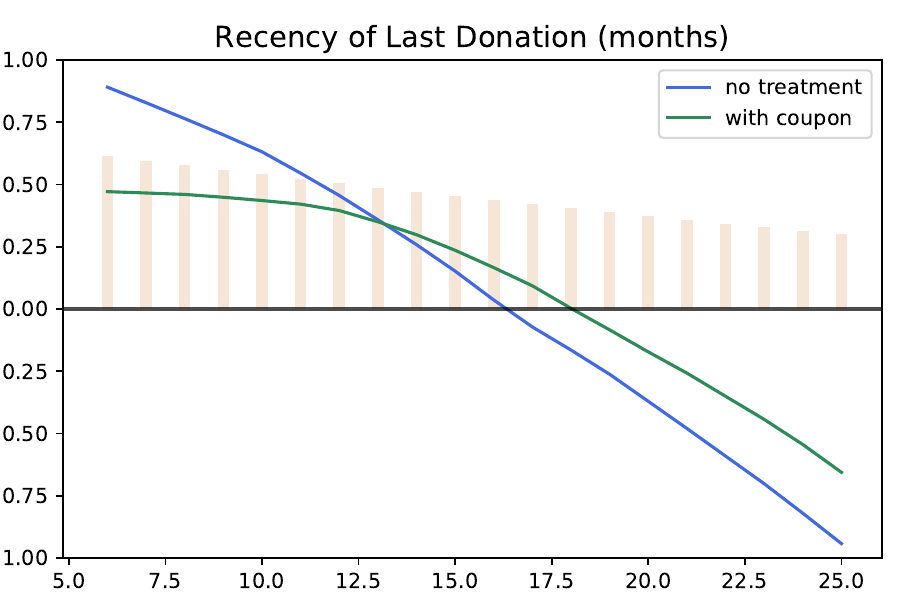}
\includegraphics[width=0.23\textwidth]{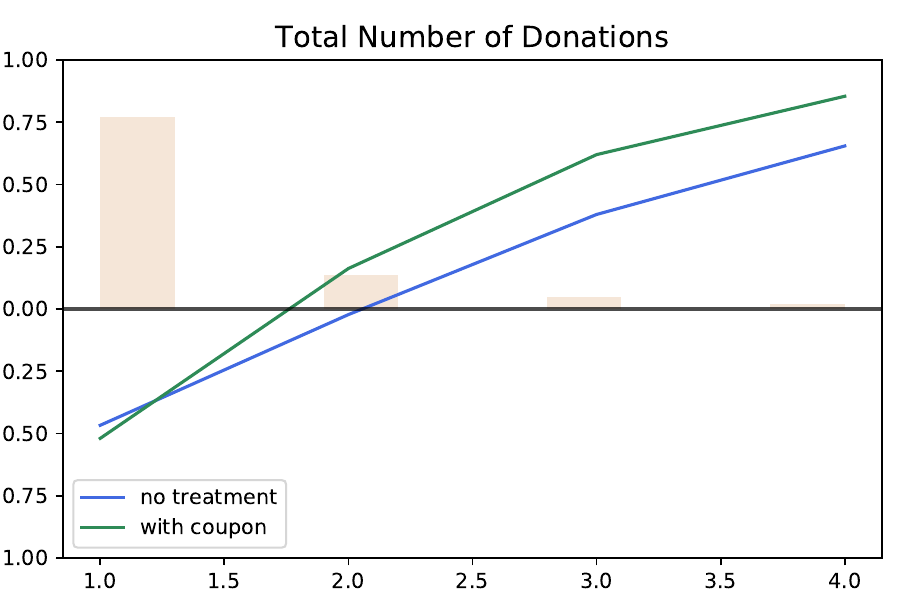}
\includegraphics[width=0.23\textwidth]{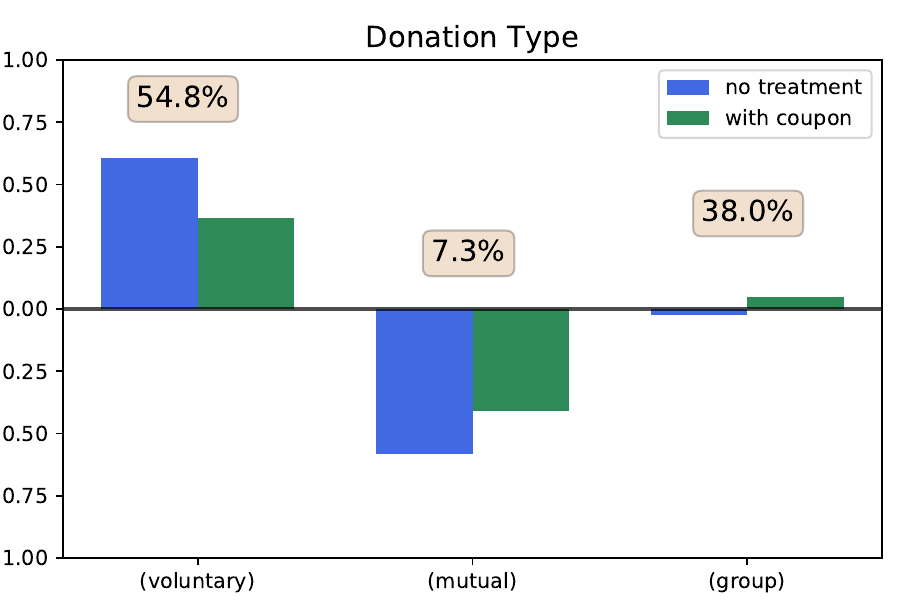}
\includegraphics[width=0.23\textwidth]{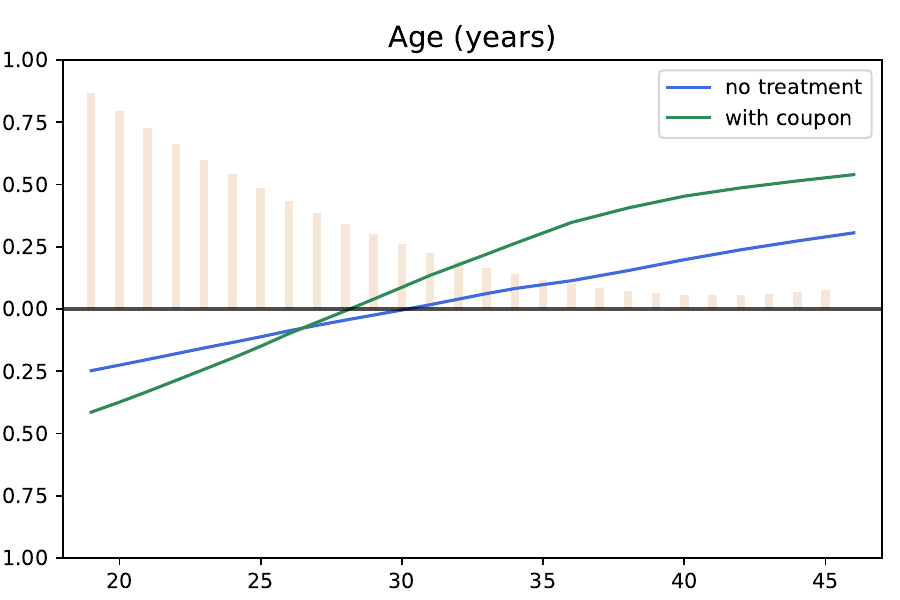}

\includegraphics[width=0.23\textwidth]{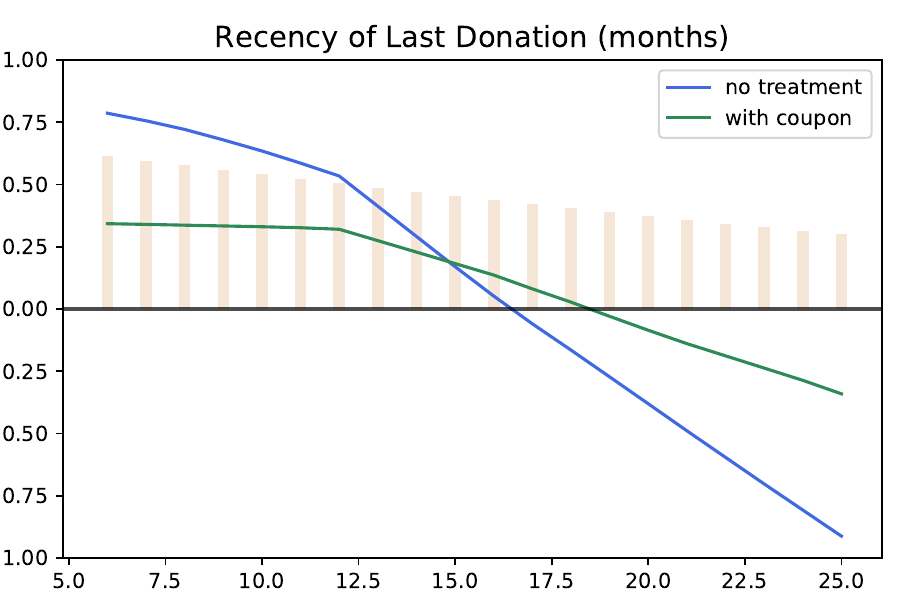}
\includegraphics[width=0.23\textwidth]{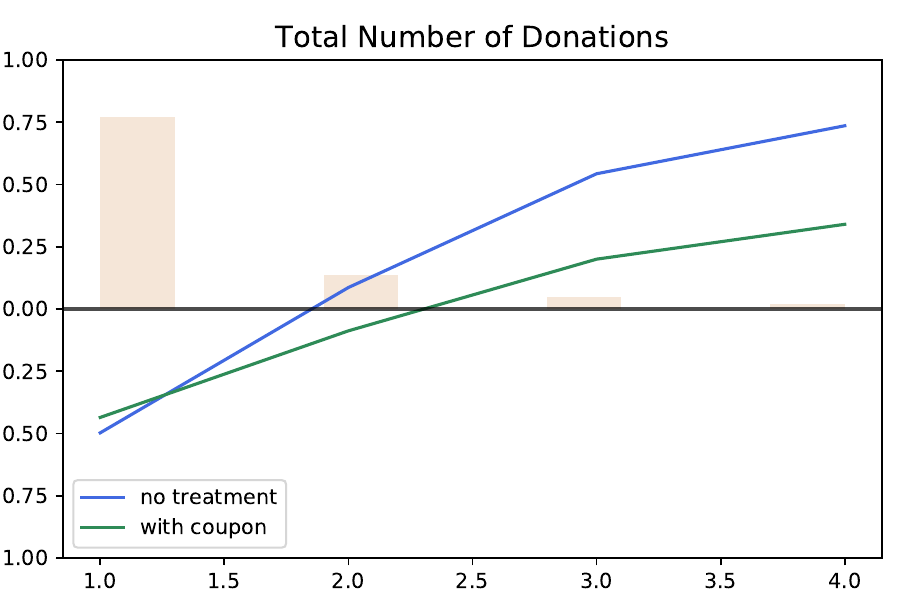}
\includegraphics[width=0.23\textwidth]{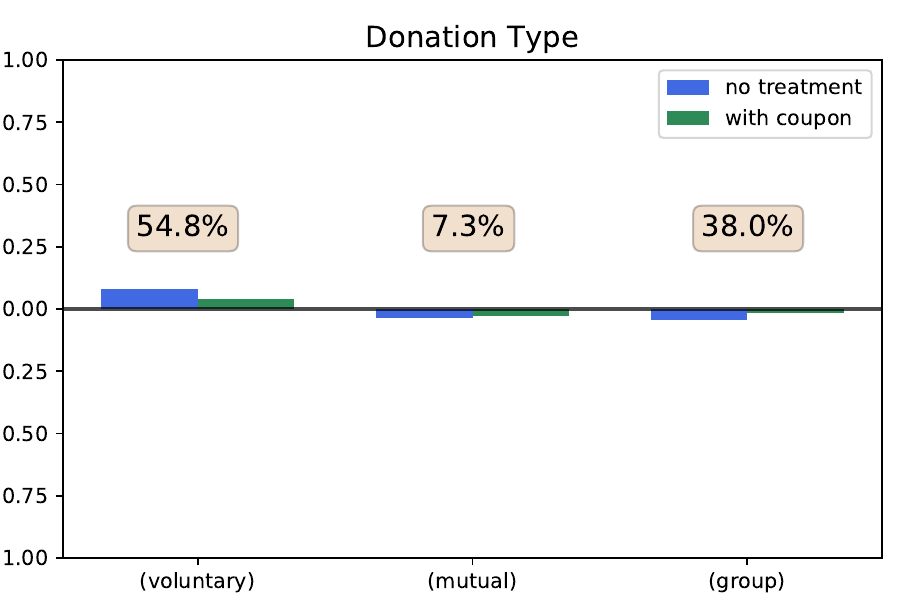}
\includegraphics[width=0.23\textwidth]{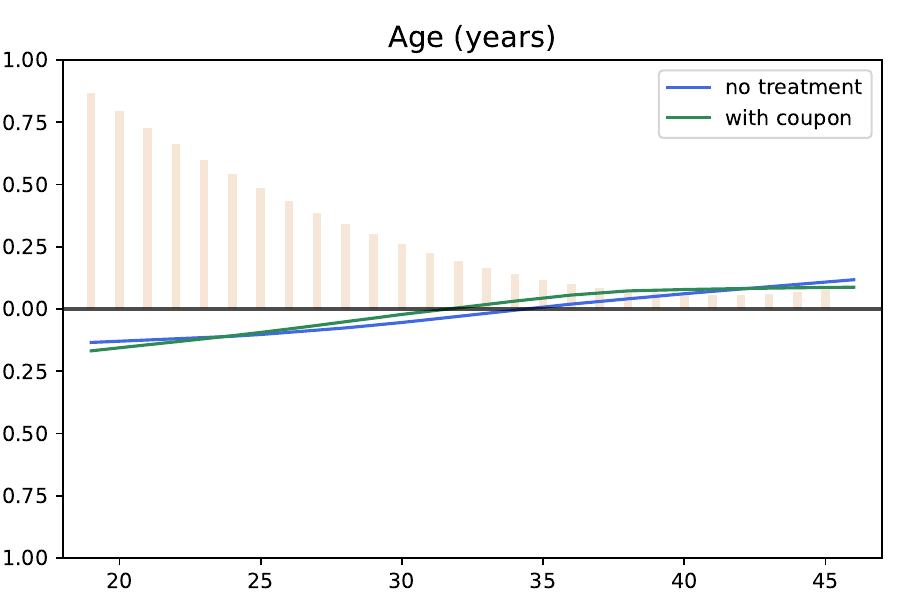}
\caption{Blood Donation 1D Likelihood Trends. 
\textnormal{The top row corresponds to the DNN trends; the bottom row corresponds to the GAM1 trends. The blue lines correspond to T=0 and the green lines correspond to T=1.  The orange shading illustrates the feature's distribution.  These top four 1D trends capture $45.8\%$, $14.8\%$, $9.7\%$, and $7.0\%$ respectively of the total variance of the DNN.}}
\label{fig:blood_1d_trends}
\end{figure*}

\begin{figure*}[h]
\centering
\includegraphics[width=0.23\textwidth]{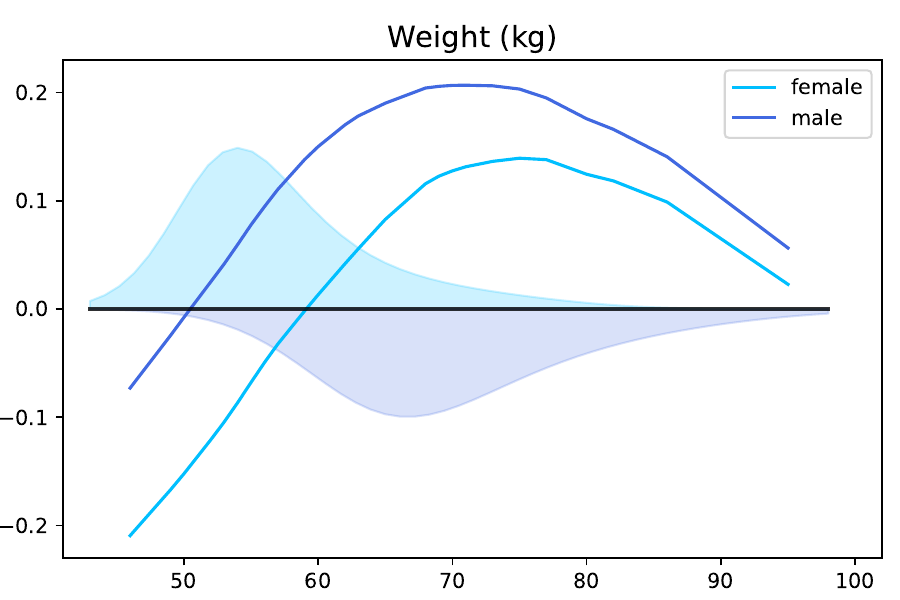}
\includegraphics[width=0.23\textwidth]{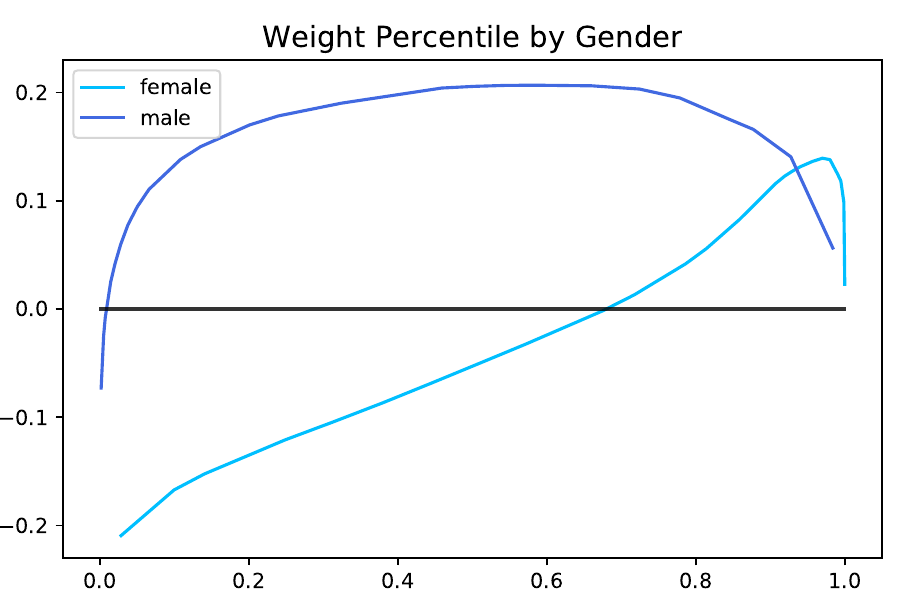}
\includegraphics[width=0.23\textwidth]{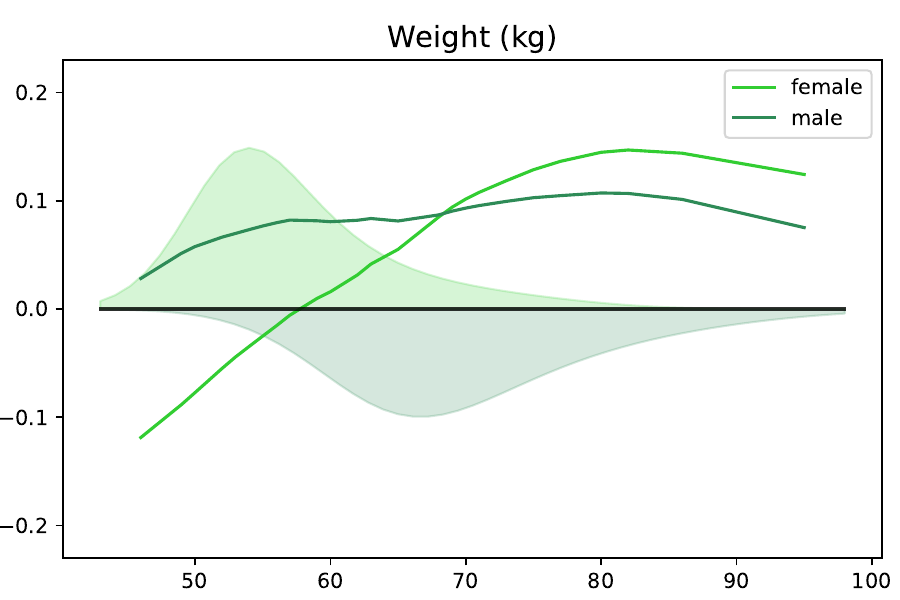}
\includegraphics[width=0.23\textwidth]{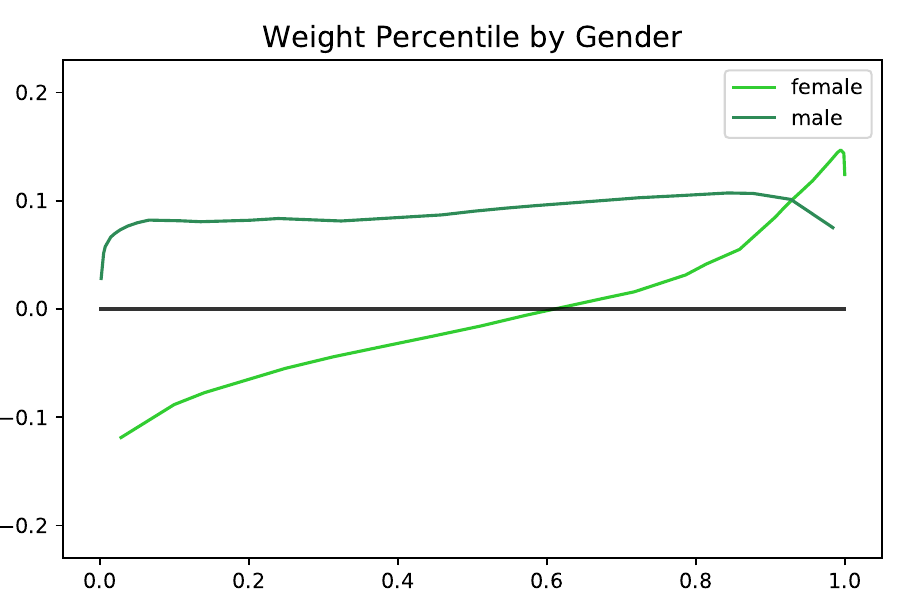}

\includegraphics[width=0.23\textwidth]{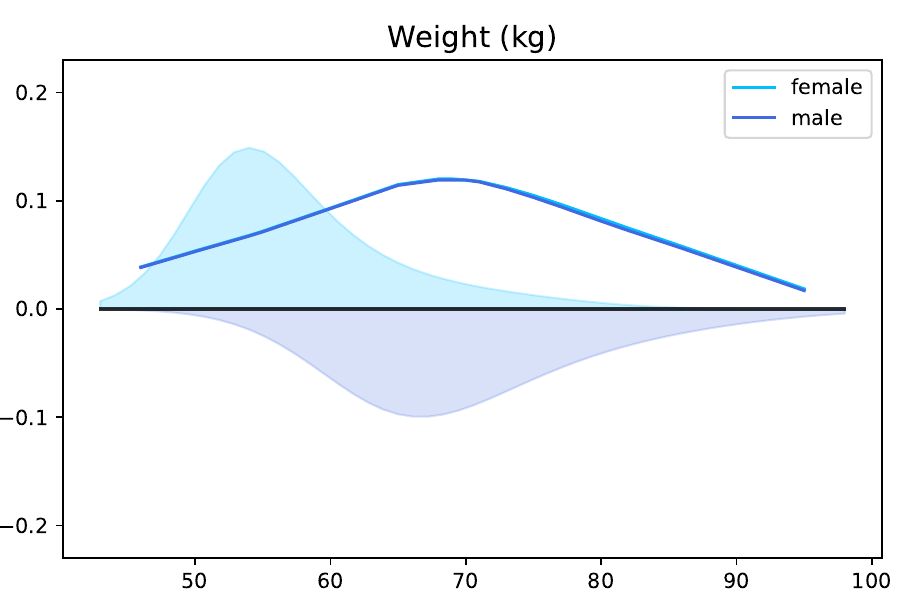}
\includegraphics[width=0.23\textwidth]{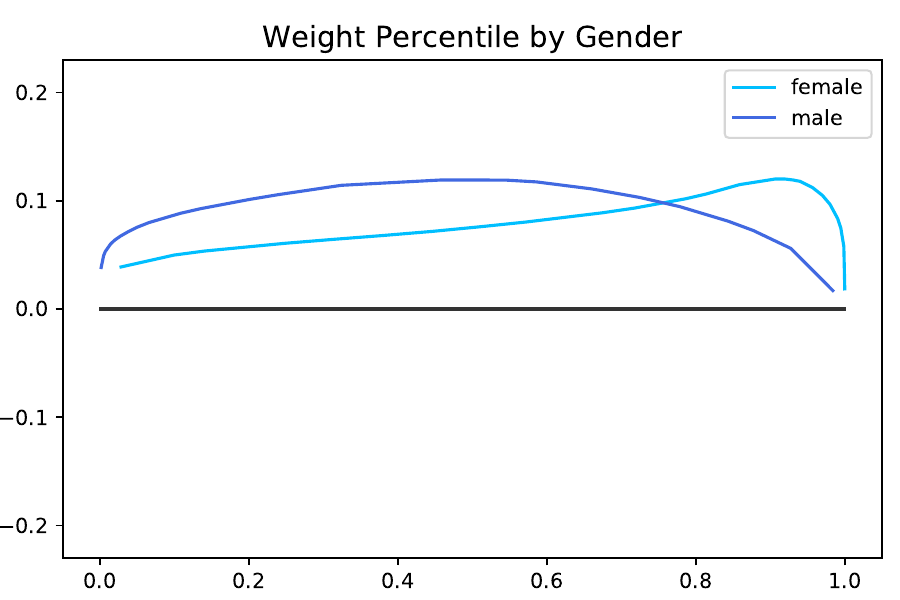}
\includegraphics[width=0.23\textwidth]{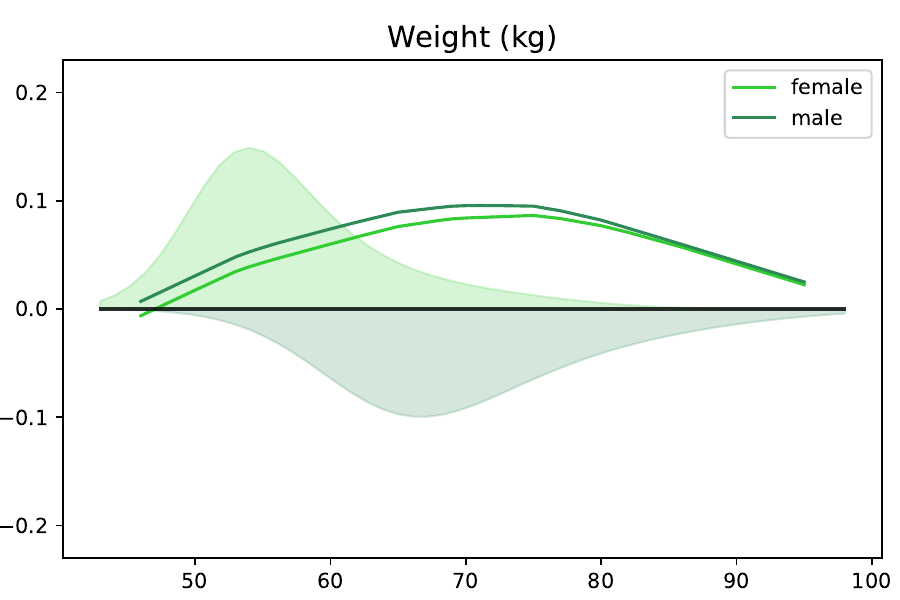}
\includegraphics[width=0.23\textwidth]{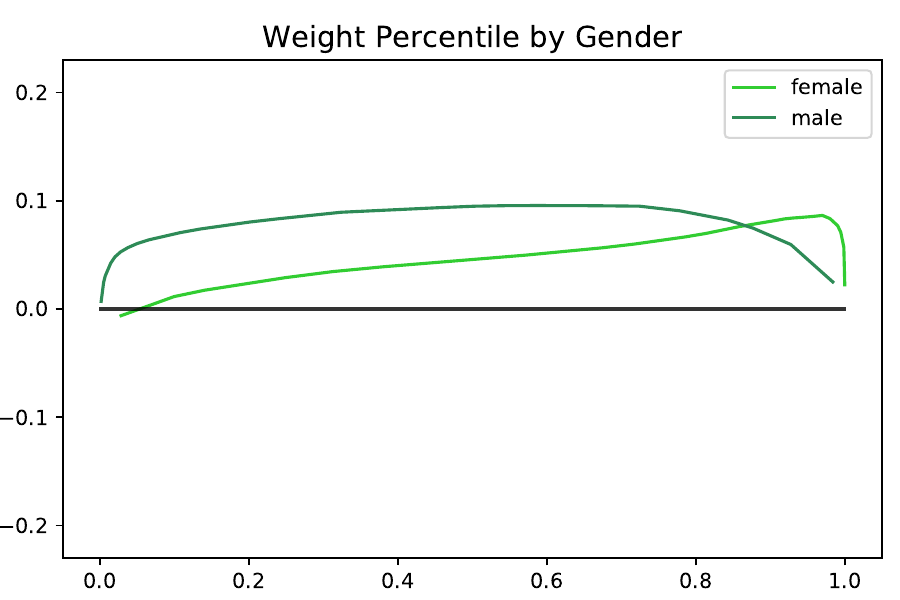}
\caption{Blood Donation 2D Likelihood Trends. 
\textnormal{The top row corresponds to the DNN trends; the bottom row corresponds to the GAM2 trends.  The left four blue figures correspond to no treatment and the right four green figures correspond to the treatment.  The background shading illustrates the distribution of weights for each gender (female above; male below.) }
}
\label{fig:blood_2d_trends_gend_weight}
\end{figure*}

\subsection{Interpretation}
After training our blackbox algorithm, we use Archipelago and our GAM model to interpret the structure of the data.
We trained one GAM to mimic the behavior of the blackbox DNN and another GAM to directly predict the result.
Both were trained on a smaller auditing set to simulate what might be practically available to an external audience.

In Figure \ref{fig:blood_1d_trends}, we first see the one-dimensional trends from the blood donation dataset.
For the DNN, a large majority (about 90\%) of the prediction variance is already captured with these 1D trends alone.
Using our learned interactions, however, the GAM2 can represent $98.5\%$ of the DNN's prediction variance.
We highlight that this means our interpretable model can essentially match the DNNs predictions almost exactly.

In Figure \ref{fig:blood_1d_trends},
we see there is mostly alignment in the first two learned shape functions (recency of last donation and total number of donations); however, there is a large discrepancy in the importance of donation type.
In particular, there are much larger discrepancies in the regions where there are fewer test examples.
The DNN consistently overestimates effects in these low-data regions like patients having more than three donations or patients having greater than forty years of age.

In Figure \ref{fig:blood_2d_trends_gend_weight}, we see the interaction between gender and weight in the model.
The differences in treatment between genders is not as pronounced until we reconsider treatment assignment as a function of weight percentile instead of raw weight.
We moreover see that the auditing GAM2 believes that weight and gender have a smaller impact on the outcome than what the DNN model believes.
It seems the DNN has amplified the effect that gender has on prediction outcome and we explore the result of toning down the exacerbated trends learned by the DNN model in the next section.

\subsection{Improvement}
\label{sec:results_improvement}

For our blood donation dataset, we consider removing sensitive shape functions in two different ways: (a) completely removing the shape function and replacing it with the baseline zero function; and (b) replacing the shape function with the GAM audited shape function.
We see in Table \ref{tab:blood_gend_and_age_removed} the result of removing the gender and weight interaction visualized above in Figure \ref{fig:blood_2d_trends_gend_weight}.
Both methods of removing the biased terms result in increased treatment fairness and both also result in slightly increased outcome fairness.
The economic benefit is relatively stable as we remove the biased term.

\begin{table}[h!]
\parbox{.48\textwidth}{
    \begin{centering}
    \caption{Blood Donation: Gender and Weight Shapes Removed}
    \small
    \scriptsize
    \resizebox{.475\textwidth}{!}{%
    \begin{tabular}{|c|c|c|c|c|c|c|c|}
    \cline{3-8}
        \multicolumn{2}{c|}{} & \multicolumn{3}{c|}{gender removed} & \multicolumn{3}{c|}{age removed}\\
    \hline 
         \multicolumn{1}{|c|}{} & percent & economic & \multicolumn{2}{c|}{gender} & economic & \multicolumn{2}{c|}{age} \\ 
         \multicolumn{1}{|c|}{} & removed & benefit (RMB) & \multicolumn{1}{c}{TF} & \multicolumn{1}{c|}{OF} & benefit & \multicolumn{1}{c}{TF} & \multicolumn{1}{c|}{OF} \\
         \hline
          DNN & $0.0\%$ & 2.98$\pm$0.14  &  $92.4$ &  $57.4$ & 2.97$\pm$0.14  &  $85.0$ &  $61.5$ \\ 
          removed & $20.0\%$ & 2.99$\pm$0.15  &  $93.9$ &  $57.9$ & 2.95$\pm$0.09  &  $89.6$ &  $62.2$ \\ 
           & $40.0\%$ & 3.01$\pm$0.17  &  $95.5$ &  $58.5$ & 2.97$\pm$0.13  &  $93.4$ &  $62.9$ \\
           & $60.0\%$ & 2.98$\pm$0.15  &  $97.0$ &  $58.1$ & 2.97$\pm$0.13  &  $97.5$ &  $60.6$ \\ 
           & $80.0\%$ & 2.98$\pm$0.16  &  $98.5$ &  $58.9$ & 2.97$\pm$0.13  &  $98.4$ &  $63.1$ \\ 
           & $100.0\%$ & 2.98$\pm$0.16  &  $98.6$ &  $59.7$ & 2.97$\pm$0.13  &  $94.6$ &  $65.6$ \\ 
         \hline
        DNN          & $0.0\%$ & 2.98$\pm$0.14  &  $92.4$ &  $57.4$ & 2.97$\pm$0.14  &  $85.0$ &  $61.5$ \\ 
        removed,  & $20.0\%$ & 2.99$\pm$0.15      &  $94.0$ &  $58.0$ & 2.93$\pm$0.13  &  $87.1$ &  $62.0$ \\ 
        GAM  & $40.0\%$ & 3.01$\pm$0.17  &  $95.6$ &  $58.6$ & 2.94$\pm$0.15  &  $89.0$ &  $63.3$ \\ 
        added   & $60.0\%$ & 2.98$\pm$0.15  &  $97.2$ &  $58.2$ & 2.97$\pm$0.11  &  $90.6$ &  $63.9$ \\ 
           & $80.0\%$ & 2.97$\pm$0.16  &  $98.4$ &  $59.1$ & 2.94$\pm$0.09  &  $92.3$ &  $64.3$ \\ 
           & $100.0\%$ & 3.00$\pm$0.12  &  $98.5$ &  $59.1$ & 2.93$\pm$0.14  &  $94.5$ &  $63.4$ \\ 
           \hline
    \end{tabular}} \\[10pt]
    \label{tab:blood_gend_and_age_removed}
    \end{centering}
    }
\end{table}

We also see in Table \ref{tab:blood_gend_and_age_removed} the effect of removing the interaction terms depending on age, including the trend visualized in Figure \ref{fig:blood_1d_trends}.
Both methods slightly improve the outcome fairness and keep the economic benefit relatively stable.
Completely removing the DNN age trends finds a sweet spot for $TF$ around $80\%$ removal, however, ultimately goes back down to $94.6\%$ treatment fairness, likely overcorrecting the original bias.
Replacing the DNN age trends with the GAM age trends instead seems to gradually improves the treatment fairness from $85.0\%$ to $94.5\%$.

We generally see that this interpretable technique of adjusting the blackbox model predictions has minimal impact on changing the economic value of a policy, while mildly improving its fairness.
Such changes can be clearly quantified and visualized by comparing the shape functions from before and after the adjustment, as in Figures \ref{fig:blood_1d_trends} and \ref{fig:blood_2d_trends_gend_weight}.
Further experiments on our Collage dataset exploring its shape functions are left to Appendix \ref{app_sec:additional_results}.

\begin{table}[h]
    \begin{centering}
    \caption{Synthetic Dataset Performance (MSE of ITE)}
    \label{tab:synthetic_measurements}
    \small
    \scriptsize
    \begin{tabular}{|c|c|c|c|c|c|}
    \hline
         & \multicolumn{5}{c|}{correlation coefficient (c)}  \\ 
         \cline{2-6}
         model & 0.00 & 0.25 & 0.50 & 0.75 & 1.00 \\ 
         \hline
DNN         & 0.011 & 0.010 & 0.009 & 0.007 & 0.007  \\ 
GAM1        & 0.063 & 0.058 & 0.060 & 0.061 & 0.069  \\ 
GAM2        & 0.007 & 0.007 & 0.006 & 0.006 & 0.006  \\  
\hline
GAM2\_{X3}     & 0.054 & 0.031 & 0.075 & 0.078 & 0.069  \\ 
GAM2\_{all-S}   & 0.055 & 0.031 & 0.076 & 0.080 & 0.078  \\ 
        \hline
    \end{tabular} \\[10pt]
    \end{centering}
\end{table}

For the synthetic dataset in Table \ref{tab:synthetic_measurements}, we indeed find that the GAM1 model struggled to model the feature interactions of our synthetic ITE function.
The GAM2 model, however, accurately predicts the correct ITE with test error similar to the DNN across all levels of correlation.
We can additionally see that after explicitly removing the shape functions corresponding the the third sensitive feature $X3$ and removing all sensitive features $S$, there is a drop in the ability to accurately predict.

\begin{figure}[h]
\centering 
\includegraphics[width=0.23\textwidth]{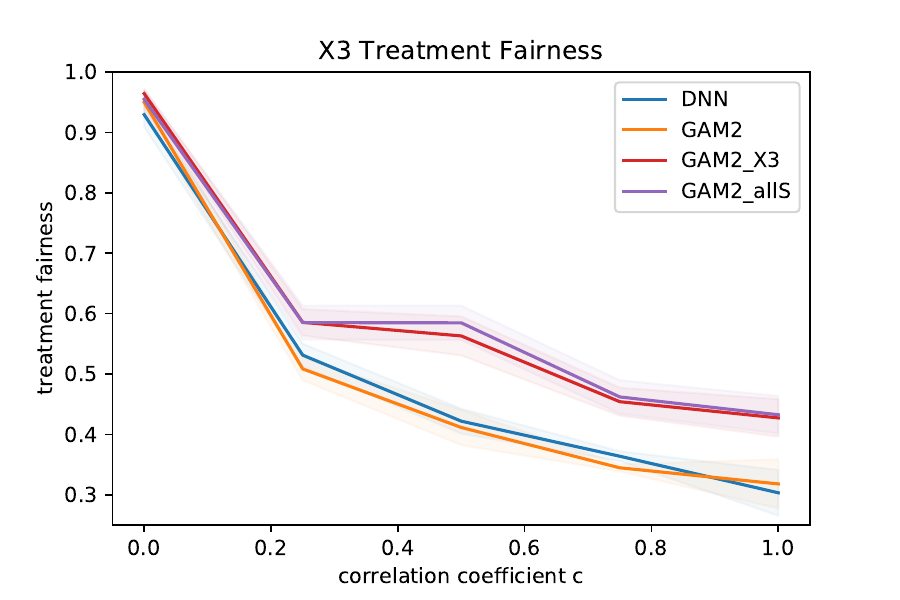}
\includegraphics[width=0.23\textwidth]{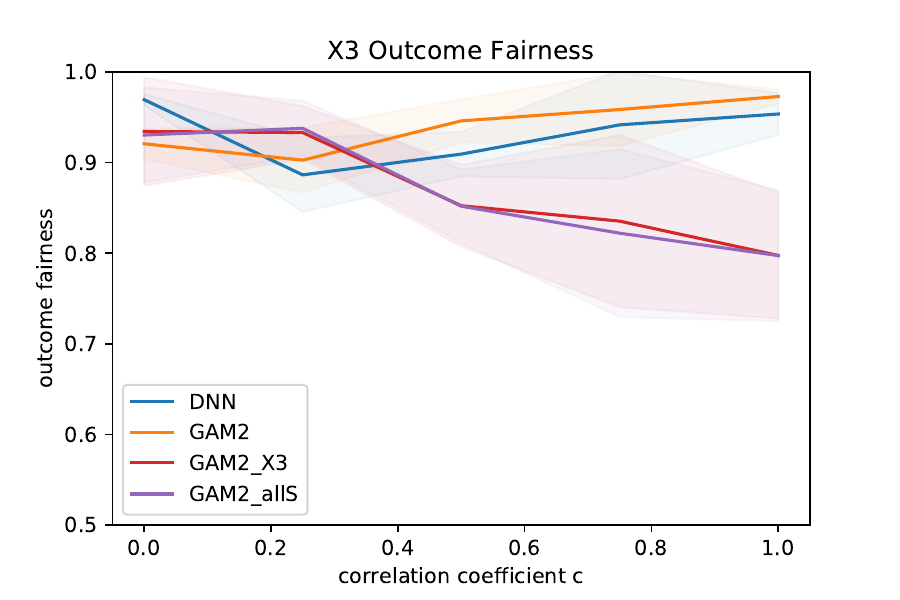}
\includegraphics[width=0.23\textwidth]{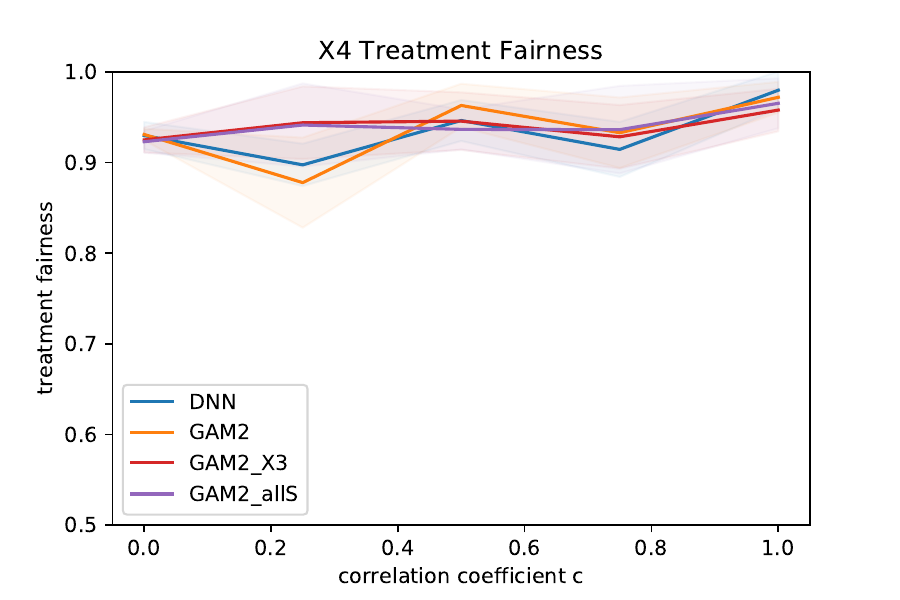}
\includegraphics[width=0.23\textwidth]{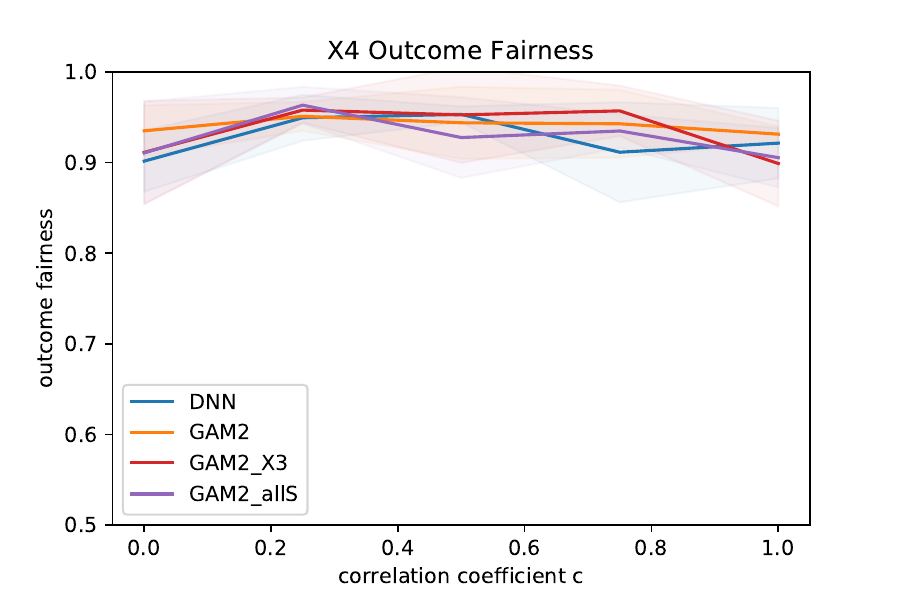}
\caption{TF and OF for the Synthetic Dataset.}
\label{fig:synth_fairness_plots}
\end{figure}

In Figure \ref{fig:synth_fairness_plots}, we can also see in the synthetic dataset, how the treatment fairness and outcome fairness change for each algorithm as we increase the correlations with sensitive features.
First, in the two right-most plots, we observe that for the sensitive feature $X_4$ which is independent from all other $X$, all models treat $X_4$ relatively the same, with values for both treatment fairness and outcome fairness consistently hovering above 90\%.
For the sensitive variable $X_3$, however, the story is much different.
We see that as we increase the correlation $X_3$ has with important prediction variables, both the DNN and GAM2 model begin to treat constituents more disparagingly with respect to feature $X_3$.
This effect on treatment fairness is lessened when we remove shape functions from the GAM2 model, however, this change also degrades the outcome fairness for $X_3$, with the GAM2 models which remove $X_3$ both dipping below 80\% outcome fairness.

\subsection{Exploring the Price of Fairness}
One of the most popular techniques from the post-processing category of fairness is the technique of multiple thresholds, creating a different treatment threshold for each protected class.
Existing studies on real-world data examine the impact these thresholds have on treatment fairness; however, in our experiments, we simultaneously investigate the outcome fairness and economic impact alongside the treatment fairness.

\begin{figure}[h]
\centering
\includegraphics[width=0.40\textwidth]{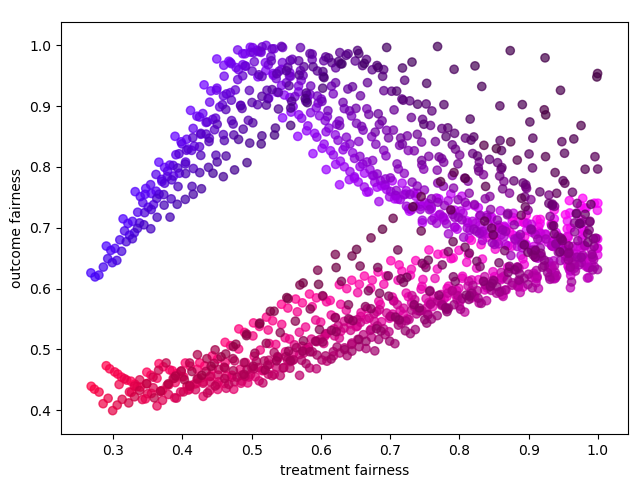}
\caption{Postprocess Thresholding for DNN Model. \textnormal{Each point represents a policy corresponding to a unique threshold, plotted with the final treatment fairness and outcome fairness it achieved.}
}
\label{fig:blood_gend_2d_fairness_noStars}
\end{figure}

\begin{figure}[h]
\centering
\includegraphics[width=0.40\textwidth]{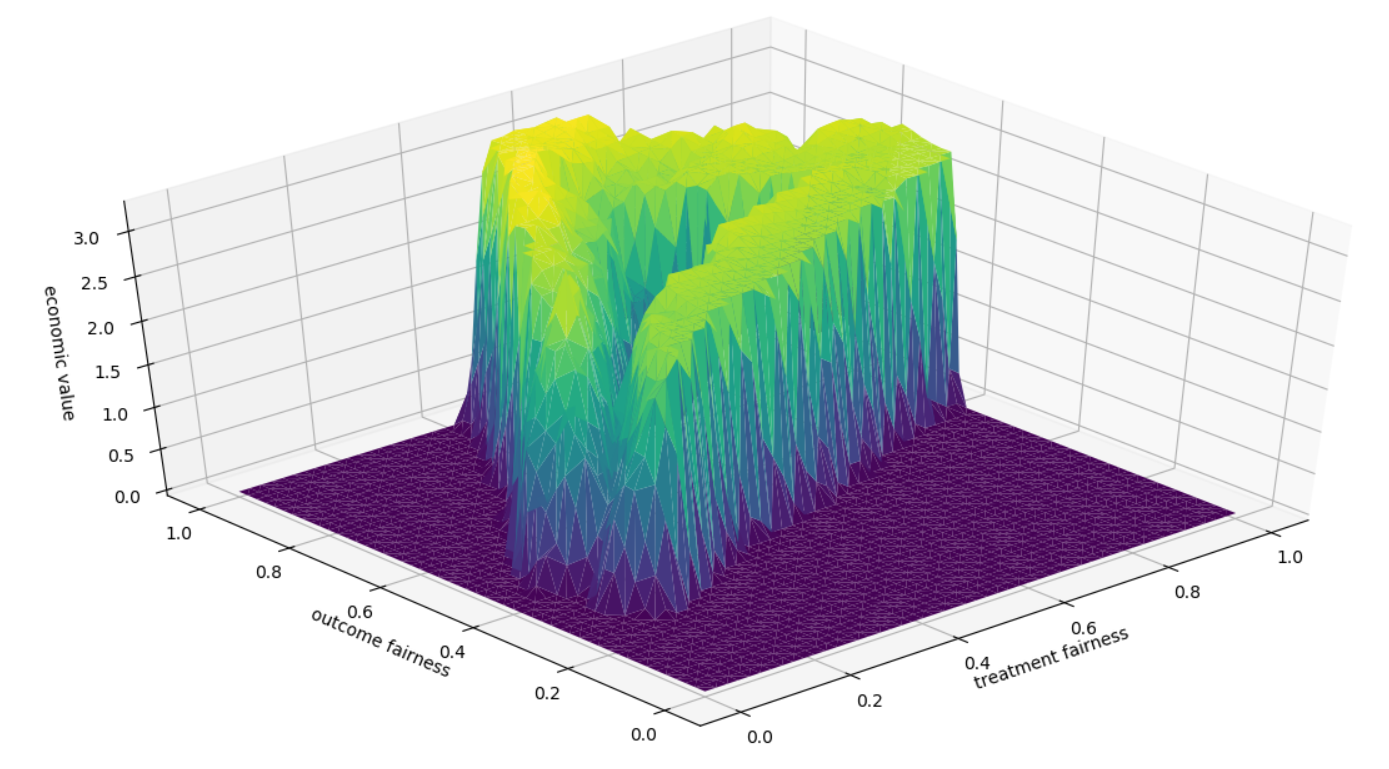}
\caption{Thresholding Manifold for a DNN Model. \textnormal{The three axes are treatment fairness, outcome fairness, and economic value. Policies cluster around triangle-like surface which obtains relatively similar economic performance.}}
\label{fig:3d_fairness_1_model}
\end{figure}

Figure \ref{fig:blood_gend_2d_fairness_noStars} depicts where many different thresholds end up on a 2D chart comparing TF against OF.
Each dot corresponds to a program evaluation using the randomized experiments.
More blue refers to a greater percentage of men being treated; more red refers to a greater percentage of women being treated.
On the right side of the figure, we see the magenta corner corresponding to $100\%$ treatment fairness, where an equal proportion of men and women receive the promotion.
In order to achieve outcome fairness, however, we need to treat a larger percentage of the male population.
Balancing the \textit{opportunity of a coupon} with the \textit{burden of donating blood} and further debating which of these is the `correct' decision is at the heart of what the fairness literature has been disputing over.
Directly visualizing this tradeoff in a practical setting is an exciting result of our setup.

Figure \ref{fig:3d_fairness_1_model} adds the third dimension of economic benefit to Figure \ref{fig:blood_gend_2d_fairness_noStars}.
For visual clarity, we replace the scatter plot with a manifold approximation, coalescing the results of multiple thresholdings.
In the far side of the corresponding surface, we see a `ridge of fairness' where we achieve similar economic benefit and are trading off between outcome fairness and treatment fairness, 
providing evidence that the technique of multiple thresholds could generally be an economic strategy to adjust for fairness.
Seemingly, multiple thresholding like all other techniques we considered is unable to achieve perfect treatment and outcome fairness simultaneously.
We find similar results across other model types in Appendix \ref{app_sec:additional_results}, providing evidence towards a consistent dataset bias rather than just an algorithmic bias.

We also briefly recall to the introductory Figure \ref{fig:college_different_metrics} which similarly plotted multiple objectives and multiple fairness metrics for the simulated college admissions example.
Earlier, we identified that NWO is both a testable and achievable metric on this simple example.
The corresponding plot in Figure \ref{fig:3d_fairness_1_model} is able to give the same insights for optimizing multiple objectives, but importantly is able to work for a real-world dataset without needing an SCM or simulation.
We imagine this gives significantly greater flexibility both to algorithm designers and to external auditors.

\subsection{Ethical Considerations of No-Worse-Off}
\label{sec:ethical_NWO}
We note that the introduced notion of no-worse-off inherently introduces a population of constituents over which we are measuring the `quality' of a decision and fails to provide individual-level guarantees, resulting in a few key sensitivities.
The first is in how to define the subpopulation.
In our college admissions example, we consider the body of rejected applicants.
However, this importantly only focuses on the students who already applied to our specific college, which may not be representative of the larger pool of college applicants.
The second issue is in how to define `worse off'.
In the same example, we assume college graduation as a sufficient definition, however, it is possible such an indicator would conceal a more nuanced outcome.
Further, group notions of fairness are always at risk of homogenizing a population and possibly aggravating discrimination within a subpopulation.
In real-world applications, we need to take care to reason about which populations we are protecting using these fairness metrics.

\section{Conclusion}
In this paper, we presented a novel framework to investigate and discuss fairness notions which may be of great practical importance to implementing fair AI methods into real-world pipelines.
We used this framework to investigate many classical fairness notions including treatment fairness, outcome fairness, post-process thresholding, and unawareness.
We developed an easily implementable distillation technique using feature interactions and GAM2 model.
We reaffirmed that in realistic scenarios, achieving both treatment fairness and outcome fairness of $100\%$ is not pragmatic, and we discovered and described both situations where economic benefit must be traded off for increased fairness and situations where it does not need to be traded away.

It is clear that this perspective on causal fairness can greatly reduce the burden of the practitioner, and future work could continue to bring additional fairness metrics and other correction procedures under this lens of causal inference.
Further development of these techniques would continue to bring fresh perspectives to older metrics and methods.
Nevertheless, {the greatest opportunity} we see to extending this work is in the direction of algorithmically random and algorithmically supervised experiments.
In order to bring causal fairness to the sensitive applications which need fairness the most, like precision healthcare and judicial decisions, it is indisputable that purely randomized experiments cannot be carried out due to ethical and moral considerations.
Accordingly, there is likely a greater need to focus on optimal experiment design and proper uncertainty estimation in order to overcome the challenges in such domains.
Success under such conditions would be widely applicable to a variety of domains and would take major steps towards advancing algorithmic fairness.



\newpage
\bibliographystyle{ACM-Reference-Format}
\bibliography{refs}

\appendix

\newpage

\section{Experiment Datasets}
\label{app_sec:datasets}
We performed our experiments on four different datasets, two synthetic datasets and two real-world datasets.
The first dataset is a simulation of college admissions for a majority group and a minority group.
The second dataset is a synthetic causal dataset where we have direct control over the correlations between the sensitive and nonsensitive variables.
The third dataset is from a Chinese blood bank sending text messages to encourage blood donations.
The fourth dataset is a marketing campaign by Collage.com attempting to expand their user base by offering rewards for inviting new customers.


\subsection{College Admission Dataset}

We use the simulation from \cite{nilforoshan2022causalConceptionsFairness} which is generated using six variables: race, education, college preparedness, test score, college decision, and college graduation.
The college is tasked with making an admission decision based on race and test score alone, knowing that race has a causal effect on educational opportunities and downstream test score, but faced with a limited budget of students to accept.
We use the parameters detailed in the appendix of their work, highlighting causal metrics which can find reasonable solutions to the college admission decision.

\subsection{Synthetic Dataset}
We created a relatively simple dataset using only 12 covariates to try to effectively study our GAM model's capacity for fitting data in the causal inference setting as well as to directly control the causal correlations and ITE function.
All 12 variables are simple mixtures of binary and Gaussian variables.  
Four of the twelve covariates are deemed as `sensitive' features $(x_1,...,x_4)$ and do not have any direct impact on the outcome; however, we attach varying levels of correlations between these features and some of the eight `nonsensitive' features $(x_5,...,x_{12})$ which do have an impact on the outcome variable.
The outcome variables are Bernoulli variables with probabilities depending on the binary treatment assignment.
These probabilities are $p_1(x) = \sigma(x_5 + x_7 +2x_9 - x_6x_7 + x_{10}x_{11})$ and $p_0(x) = \sigma(x_5 + 0.1x_8^2 +x_5x_7 - x_9x_{11})$, where $\sigma$ denotes the sigmoid function.
A simple diagram of the data generation process is shown in the figure below.
Each of the correlated variables have a correlation coefficient of $c\in \{0,\frac{1}{4},\frac{1}{2},\frac{3}{4},1\}$.

\begin{figure}[h]
    \centering
    \includegraphics[width=0.30\textwidth]{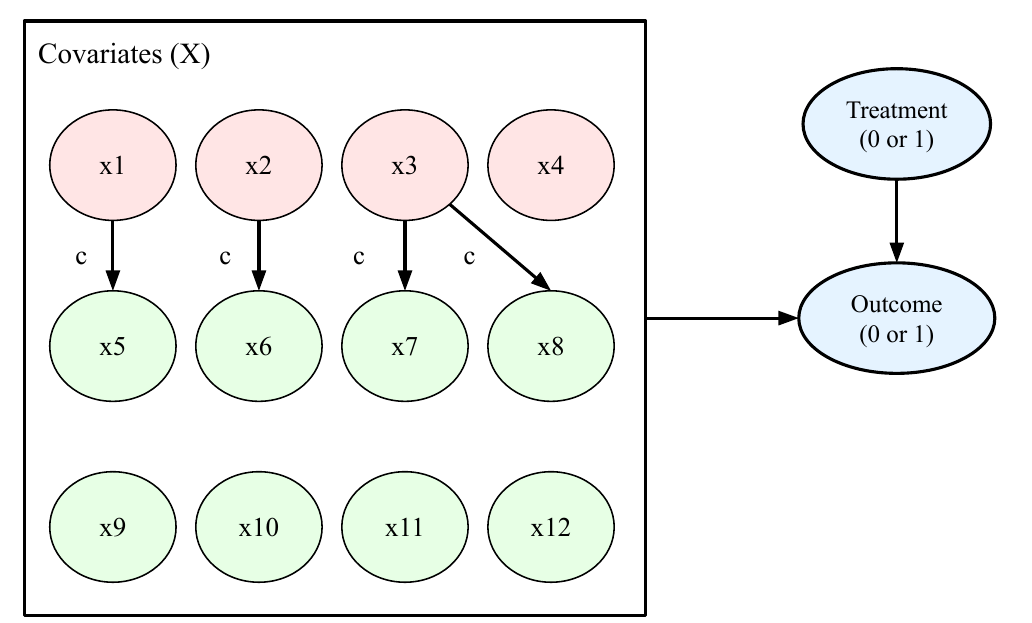}
    \caption{Diagram of Synthetic Experiment Dataset (red denotes sensitive features, green denotes nonsensitive features, arrows correspond to correlations with coefficient c)}
    \label{app_fig:synthetic_data_causal_diagram}
\end{figure}

\subsection{Blood Donation Dataset}
This dataset was collected by a blood bank whose goal was to determine the efficacy of mobile messages in targeting individuals for donation.
The treatment corresponds to receiving an invitation to donate blood in exchange for a small grocery coupon.
The members of the baseline receive no such message.
The outcome is whether the individual chooses to donate and how much blood (200/300/400ml) the individual chooses to donate.
The covariates of the individual correspond to a number of sensitive features like  gender and age as well as nonsensitive features like blood type and donation history.
In total there are 34 covariates for 60,000 potential donors.
The model attempts to predict the likelihood of an individual to donate in both cases and the ITE becomes the difference in donation likelihood.
An economic evaluation is done by bringing all factors of the experiment to the same scale as follows:
200ml of blood is evaluated at 220 RMB; the coupon gift is worth around 40 RMB; and each message costs 1 RMB to send.
Greater details on the blood bank's economic evaluation can be found in \cite{McFowland2020optimal}.
The protected attributes we explore in this dataset are gender (male/ female) and age (29-/ 30+).

\subsection{Collage Scrapbook Dataset}
This dataset was collected by the online retailer Collage.com who attempted to find the cost effectiveness of a referral program.
The treatment corresponds to sending an individual an email offering them a scrapbooking gift for every referral they make.
The outcome is whether the individual refers a friend, whether they redeem their gift, and how much their friend spends on the new platform.
The covariates of each individual again correspond to sensitive features (gender, age) and nonsensitive features (purchase history).
Overall there are 10 different features for 100,000 users.
Again, we are trying to predict the likelihood of a customer referring more customers after receiving an incentive.
The average economic gain per referral is around $\$3.93$ when no gift is given and $\$3.72$ when offered a gift.
These values are based on likelihoods to register and redeem; more details can be found in \cite{McFowland2020optimal}.
We again consider the protected attributes of gender and age.
In this setting, we find it is less meaningful to consider outcome fairness because the `exploited' party is no longer the treated individual but the friend who was recommended, hence we only consider $TF$.

\section{Feature Interactions} \label{app_sec:feature_interactions}

\noindent \textcolor{black}{\textbf{Notations}: Vectors  are represented by boldface lowercase letters, such as ${x}$ or ${w}$. The $i$-th entry of a vector ${x}$ is denoted by $x_i$. For a set $\mathcal{I}$, its cardinality is denoted by $|{\mathcal{I}}|$.  
Let $n$ be the number of features 
 in a dataset.
An \emph{interaction}, $\cI$, is the indices of a feature subset: $\cI=\{i_1,\dots,i_{|\mathcal{I}|}\}\subseteq [d] :=\{1,2,\dots,d\}$. A higher-order interaction always has $|{\cI}|\geq3$.
 For a vector ${x}\in \bbR^d$, 
  let ${x}_{\cI}\in \bbR^{|{\cI}|}$ be  restricted to the dimensions of  ${x}$ specified by $\cI$.
  Let a blackbox model be $f: \bbR^d \to \bbR$. 
}

\begin{figure*}[t]
\centering
\includegraphics[width=0.23\textwidth]{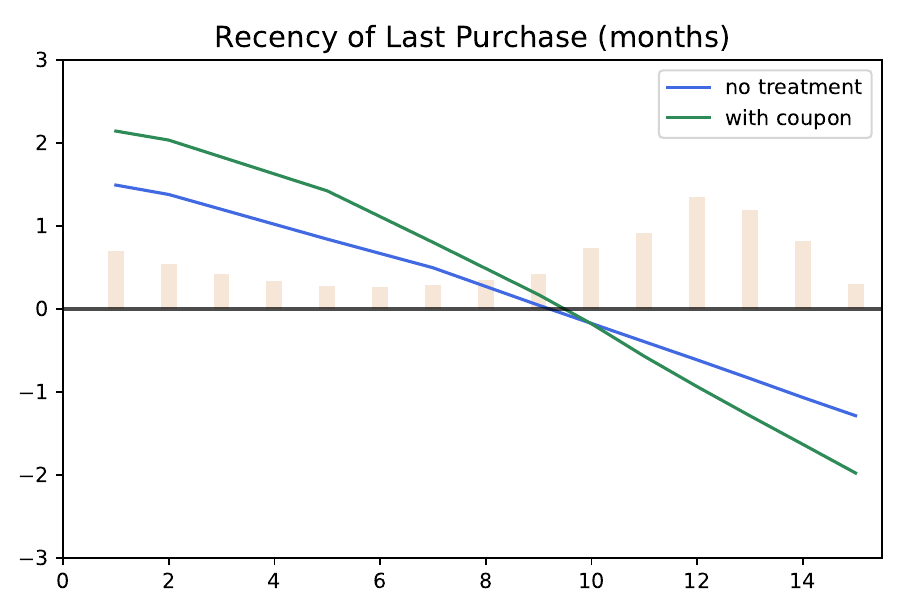}
\includegraphics[width=0.23\textwidth]{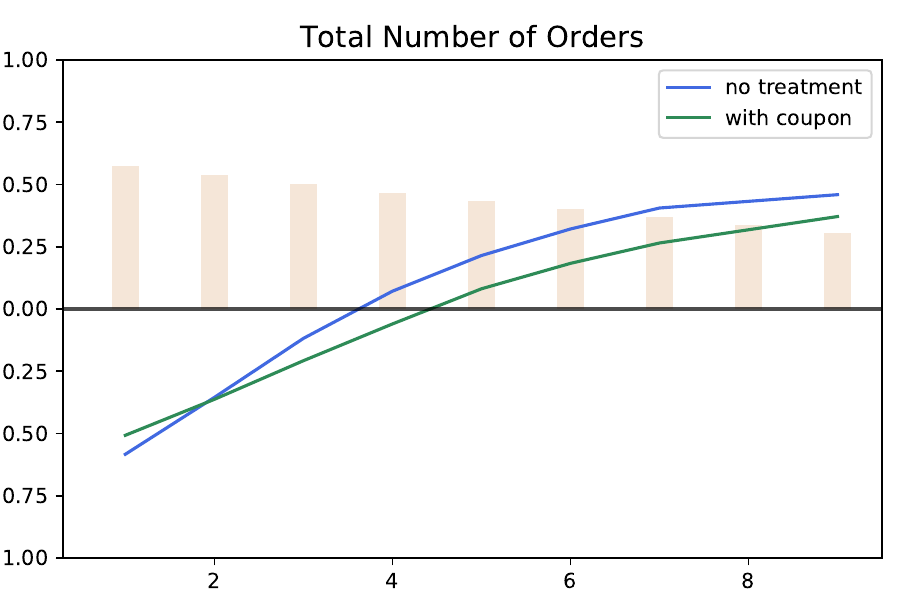}
\includegraphics[width=0.23\textwidth]{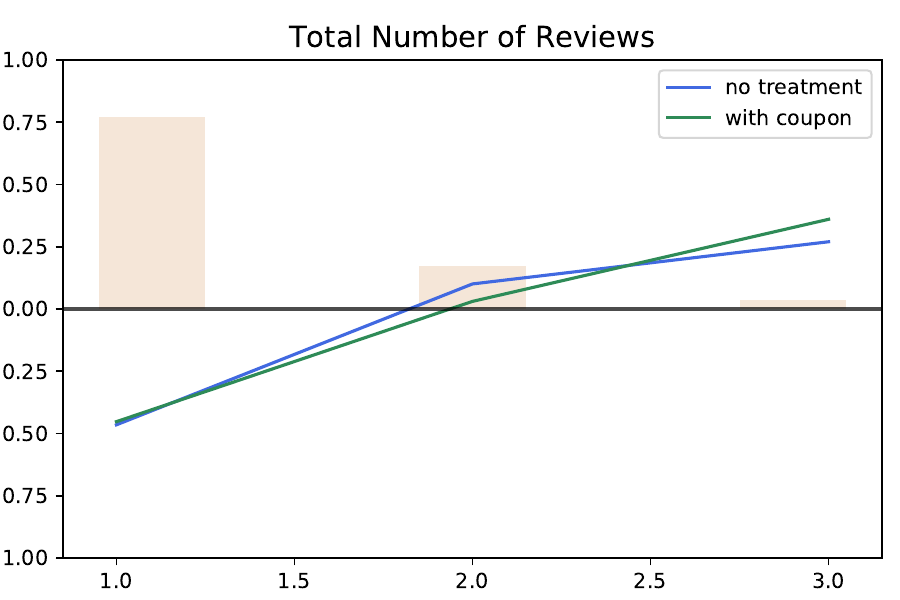}

\includegraphics[width=0.23\textwidth]{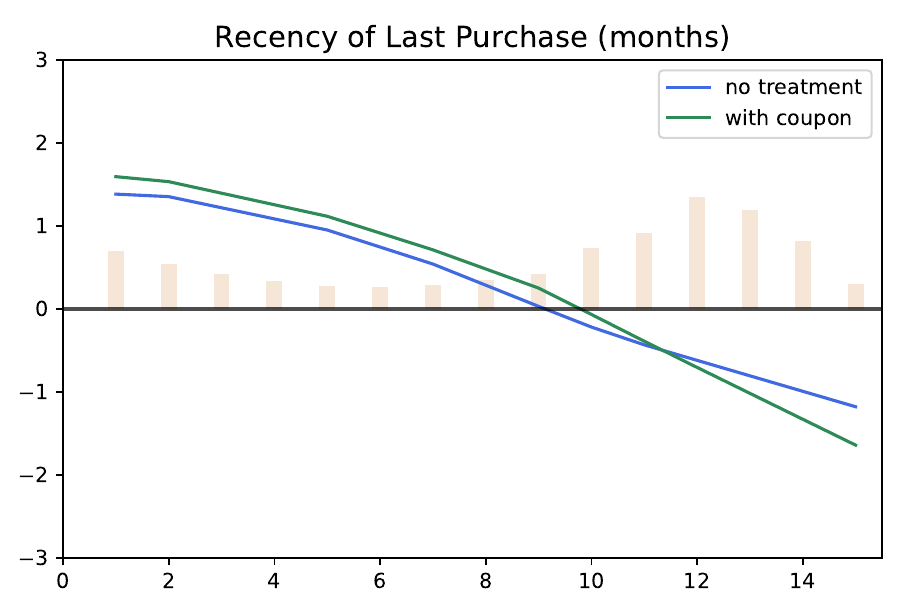}
\includegraphics[width=0.23\textwidth]{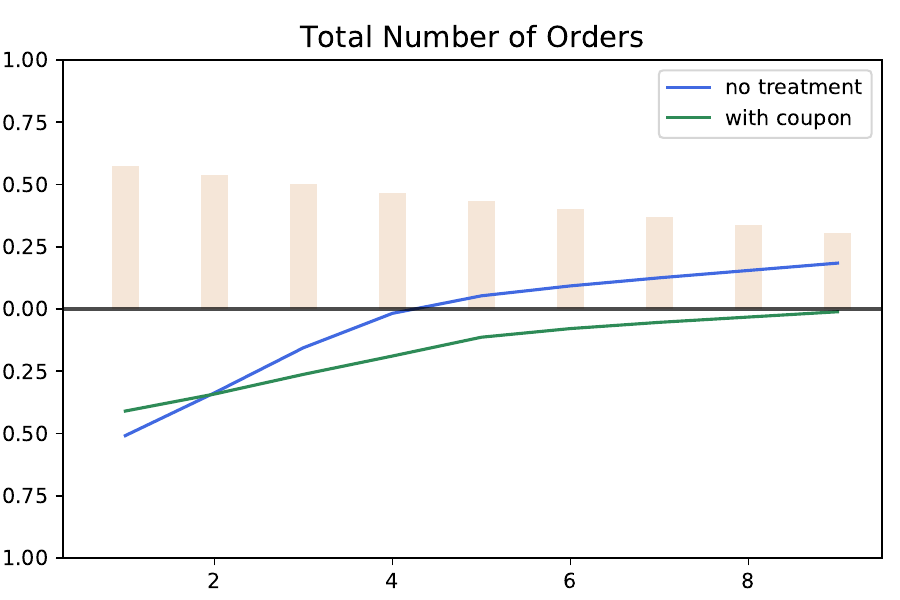}
\includegraphics[width=0.23\textwidth]{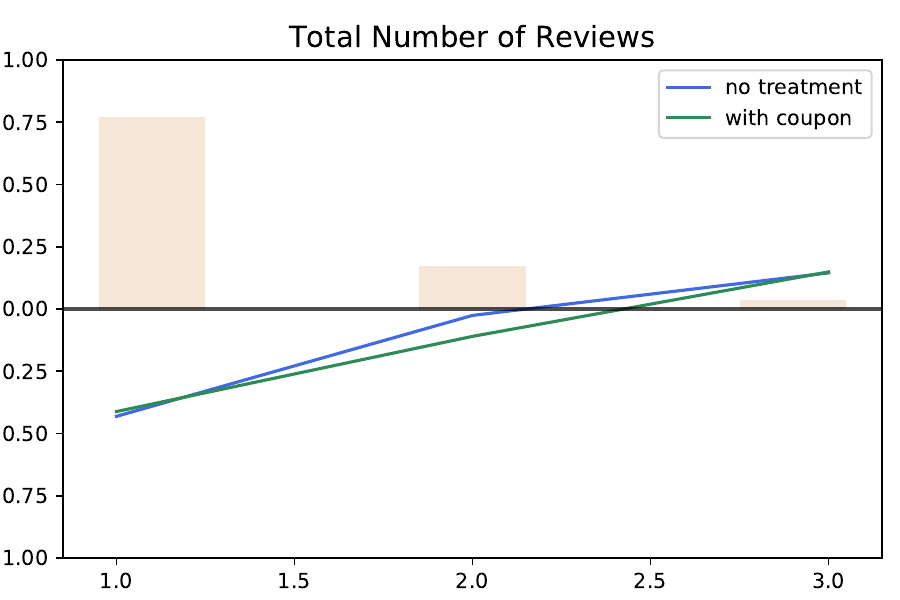}
\caption{Collage.com Referral 1D Likelihood Trends: \textnormal{The top row corresponds to the 1D likelihood trends of the DNN; the bottom row corresponds to the 1D trends learned by the GAM1.  The blue lines correspond to the trend for participants with no coupon and the green lines correspond to participants who were offered a coupon.  The orange shading in the background illustrates the approximate distribution of the given variable.  These top three 1D trends capture $82.1\%$, $6.1\%$, and $4.7\%$ respectively of the total variance of the DNN.  We see that mostly similar trends are learned by both of the models.}
}
\label{fig:collage_1d_trends}
\end{figure*}

\subsubsection{Feature Interactions}
The goal of feature interaction detection is to uncover the groups of features which depend on one another.
This is to find the sets $\mathcal{I}\subseteq[d]$ such that $\omega(\mathcal{I})$ are positive and large, where
\begin{equation}
\omega({\mathcal{I}}) := 
\bbE_{{x}}\left[\frac{\partial^{|{\mathcal{I}|}} f({x})}{\partial _{i_1}\partial x_{i_2} \dots\partial x_{i_{|\mathcal{I}|}}}\right]^2 > 0.
\label{eq:interaction_strength}
\end{equation}
This partial derivative equation corresponds to a `statistical (non-additive) feature interaction' where a function $f$ of an $n$-dimensional input $x$ cannot be decomposed into a sum of $|\cI|$ arbitrary subfunctions $f_i$ where each excludes a corresponding interaction variable~\cite{friedman2008predictive, sorokina2008detecting, tsang2017detecting}:
$f({x}) \neq \sum_{i\in\cI} f_i({x}_{\{1, 2 , \dots, n\}\setminus i}).$ 
This implies that the feature set \{$x_i$ for $i\in\cI$\} must all be simultaneously known to be able to predict the output $f(x)$.
For example, $f(x,y) = xy$ has a feature interaction whereas $f(x,y) = \log(xy)$ does not because it can be written as $\log(x) + \log(y)$.


\subsubsection{Archipelago}
Recent work with Archipelago \cite{tsang2020archipelago} approximates the Hessian of the function/ model and can very quickly detect feature interactions for any type of model given a target data instance $x^*$ and a baseline data instance $x'$.
The secant approximation of the Hessian which is used by Archipelago is defined:
{
\setlength{\abovedisplayskip}{-1.1pt}
\setlength{\belowdisplayskip}{-1.4pt}
\begin{dmath}
\scriptstyle 
    \omega_{i,j}({x})= 
    \left(\tfrac{1}{h_i h_j}
    \scriptstyle
    \left(f({x}^{\star}_{\{i,j\}} +  {x}_{\setminus{\{i,j\}}}) - f({x}'_{\{i\}} + {x}^{\star}_{\{j\}} +  {x}_{\setminus{\{i,j\}}}) - f({x}^{\star}_{\{i\}} + {x}'_{\{j\}} +  {x}_{\setminus{\{i,j\}}}) + f({x}'_{\{i,j\}} +   {x}_{\setminus{\{i,j\}}})\right)\right)^2,\label{eq:pairwise}
\end{dmath}
}

\noindent
\newline
where $h_i=x_i^*-x'_i$, $h_j=x_j^*-x'_j$,
$x^*$ is a target data instance (like a member of our validation population) and $x'$ is a baseline data instance (either the all zeroes vector or another validation sample).

The most thorough way to identify this feature interaction is using many different `context vectors' $x$.
Because of this, we theoretically have
$
\Bar{\omega}_{i,j} = \bbE_{x\in\mathcal{X}} [\omega_{i,j}(x)]
$
where the expectation is over the `full context' of all possible combinations of the target  $x^*$ and the baseline $x'$.
In this work, we only consider the approximation used by \cite{tsang2020archipelago} which greatly lessens the computation: 
{
\begin{dmath}
\Bar{\omega}_{i,j} = \frac{1}{2}(\omega_{i,j}(x^*)+\omega_{i,j}(x'))
\end{dmath}
}

This generates a score $\Bar{\omega}_{i,j}$ for every possible feature pair $\{(i,j):i\neq j\in[d]\}$.
We then rank and use the top K feature pairs for our GAM2 model which we will next describe.

\subsubsection{Generalized Additive Models} 
We consider the generalized additive model (GAM), a generalization of linear regression \cite{hastie1990originalGAM}.
We adapt the original definition to also model arbitrary interactions via 
\begin{equation}\label{eq:GAM_K}
g(y) = f_\emptyset + \sum_i f_i(x_i) +
        \sum_{k} f_{\mathcal{I}_k}(x_{\mathcal{I}_k})
\end{equation}
where $g$ represents a `link function' which in the binary classification setting will be the inverse-sigmoid and $f_\emptyset$ will be a normalizing constant.
The first two terms refer to the classical GAM, where the features do not depend on one another, but possibly have a nonlinear relationship with the output.
The third term extends GAMs to full capacity models which can represent nonlinear dependencies of arbitrary feature sets.
If our set of interactions $\{\mathcal{I}_k\}_{k=1}^K$ includes the complete set $[d]$, then our model has the same capacity as whatever underlying function model we choose for $f_i$ and $f_{\mathcal{I}}$ (splines, random forests, deep neural networks, etc.)




One of the main issues from an interpretability perspective becomes visualizing an arbitrary multivariate function for three or more dimensions.
For this reason, this paper will almost exclusively consider interactions of size two ($\mathcal{I}=\{i,j\}$) and will refer to the corresponding model as GAM2:
\begin{equation}\label{eq:GAM2_app}
g(y) = \beta_0 + \sum_i f_i(x_i) + \sum_{i<j} f_{i,j}(x_i,x_j)
\end{equation}
We will similarly refer to the original formulation with only the first two terms and no interactions as GAM1.

\begin{figure*}[h!]
\centering
\includegraphics[width=0.32\textwidth]{samples/images/dnn_vol_manif.PNG}
\includegraphics[width=0.32\textwidth]{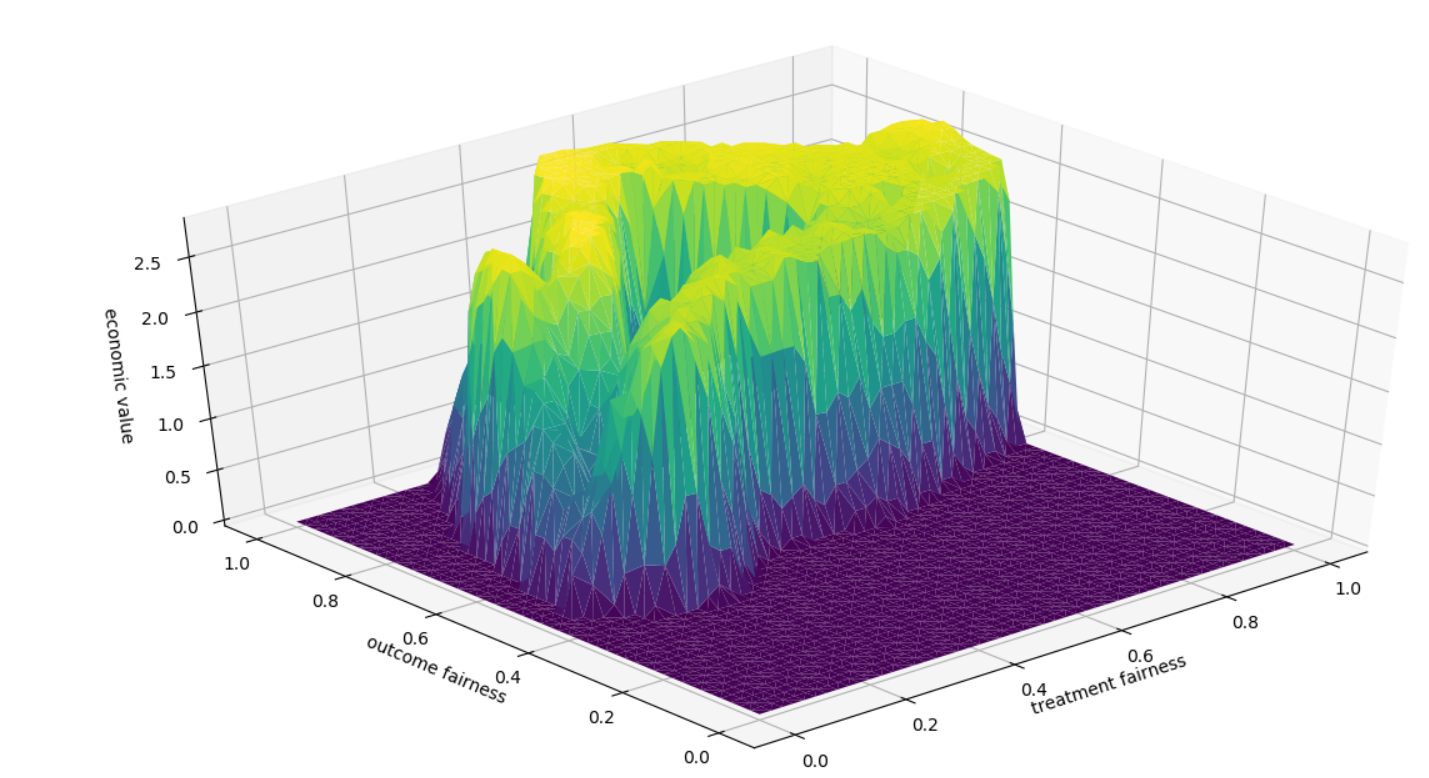}
\includegraphics[width=0.29\textwidth]{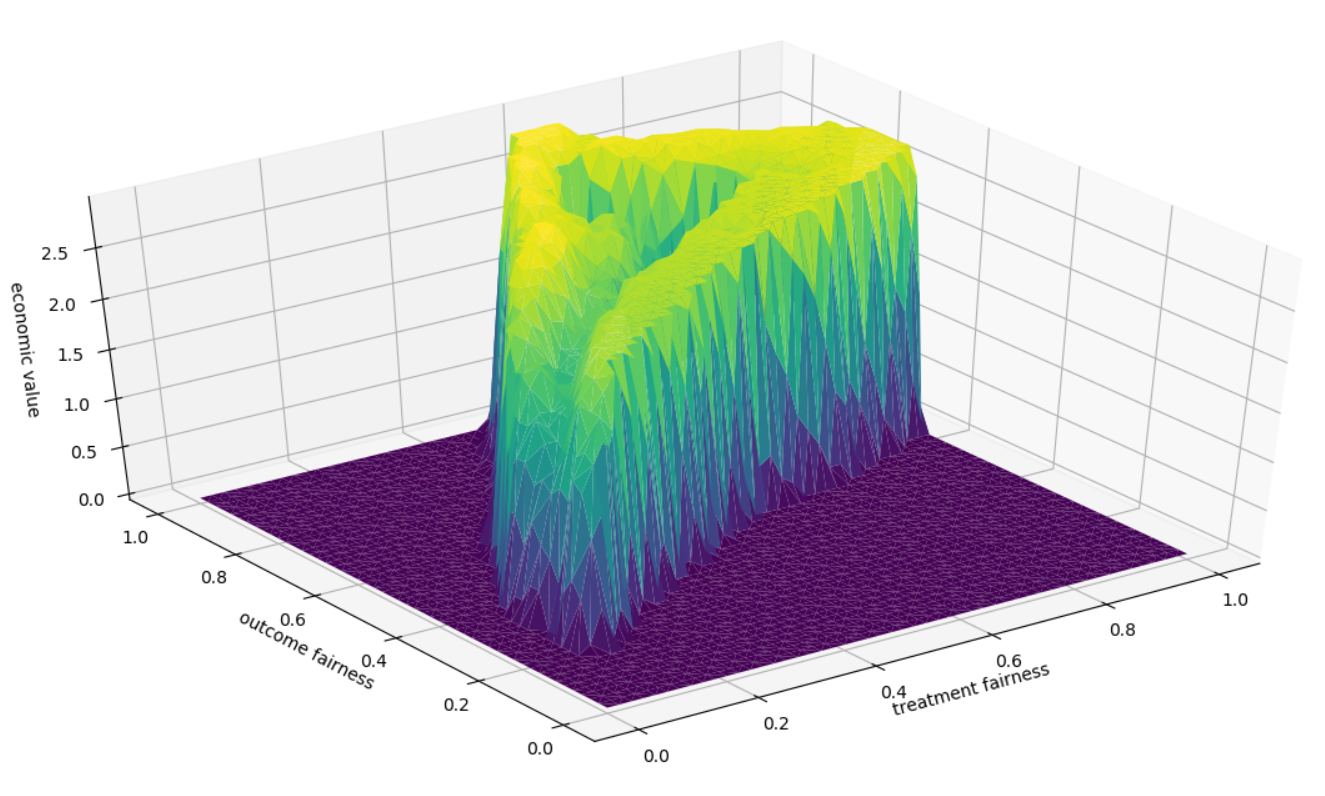}
\caption{Blood Donation: Depiction of Thresholding Manifolds for a DNN, an SVM, and an RF model (left to right)}
\label{fig:3d_fairness_3_models}
\end{figure*}

These ``shape functions'' $f_i$ and $f_{i,j}$, as they are called, can in theory be arbitrary functions.
In practice, we need a specific machine learning model which we will fit to the available data.
In our case, we use a smaller neural network for each function and fit all of these shape function networks simultaneously using AdaGrad \cite{duchi2011adagrad}.
Since we are fitting these networks simultaneously instead of using residual methods, it is also possible to view this GAM array of smaller networks as one large neural network with a very specific connection structure.

A key concern in using neural networks to fit the shape functions is keeping the number of networks low enough that our training computation time is kept reasonable.
While this is not a very large problem for the GAM1s, this quickly becomes a problem for the GAM2s.
For instance, if we have an input variable x with 34 covariates like our main dataset, then the pairwise functions would in theory need to cover all ${34\choose 2}=561$ possible pairs, corresponding to training around 600 deep networks simultaneously.
In an effort to combat this growing complexity, we use the aforementioned methods of feature interaction detection to drastically reduce the number of pairs.
We find that this model is able to maintain high quality performance with a much greater level of interpretability as we will show in our results.

\section{Additional Results} \label{app_sec:additional_results}

For the synthetic dataset, we can see that the GAM1 model struggles to model the feature interactions of our synthetic ITE model.
We see that the GAM2, however, is able to accurately model the synthetic ITE with similar test error to the DNN model across all levels of correlation.
We also include the scores of the GAM2\_X3 which removes all $X_3$ dependent shapes and the GAM2\_allS which removes all shapes which depend on any of the sensitive features in our synthetic dataset.
We see there is a large drop in MSE performance resulting from these changes and in Section \ref{sec:results_improvement} we further discuss the implications these changes have on the metrics for fairness.

\begin{table}[h]
    \begin{centering}
    \caption{Synthetic Dataset Measurement (MSE of ITE)}
    \label{app_tab:synthetic_measurements}
    \small
    \scriptsize
    \begin{tabular}{|c|c|c|c|c|c|}
    \hline
         & \multicolumn{5}{c|}{correlation coefficient (c)}  \\ 
         \cline{2-6}
         model & 0.00 & 0.25 & 0.50 & 0.75 & 1.00 \\ 
         \hline
DNN         & 0.011 & 0.010 & 0.009 & 0.007 & 0.007  \\ 
GAM1        & 0.063 & 0.058 & 0.060 & 0.061 & 0.069  \\ 
GAM2        & 0.007 & 0.007 & 0.006 & 0.006 & 0.006  \\  
\hline
GAM2_{X3}     & 0.054 & 0.031 & 0.075 & 0.078 & 0.069  \\ 
GAM2_{allS}   & 0.055 & 0.031 & 0.076 & 0.080 & 0.078  \\ 
        \hline
    \end{tabular} \\[10pt]
    \end{centering}
\end{table}

In Figure \ref{fig:collage_1d_trends}, we see the trends learned for the Collage.com dataset.
We again see that these simple and interpretable trendlines can account for a large portion of the blackbox model's prediction.
Here, in contrast to the blood donation dataset, we see that the DNN for the collage dataset learns trends which are extremely similar to the GAM trained for auditing the results.
It is likely possible this is due to the smaller dimension of this dataset preventing the DNN from focusing on the spurious correlations of higher dimensions.

\begin{figure}[h!]
\centering
\includegraphics[width=0.42\textwidth]{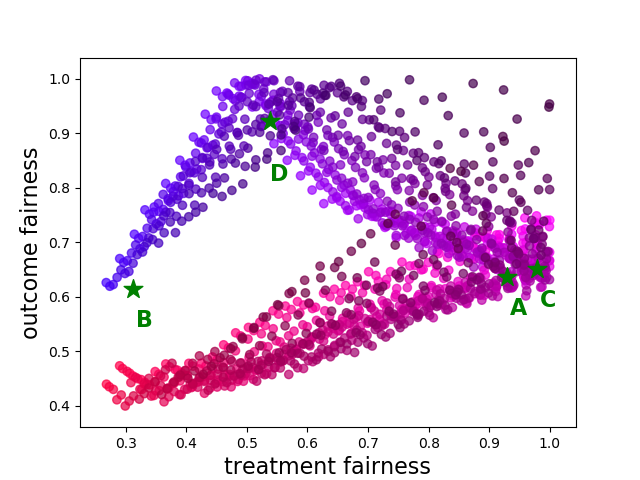}
\caption{Blood Donation: Fairness Depiction of Postprocess Thresholding for DNN Model}
\label{app_fig:blood_gend_2d_fairness_stars}
\end{figure}

\subsection{Price of Fairness}

\begin{table}[h]
    \begin{centering}
    \caption{Blood Donation Thresholding: Optimizing for Different Objectives}
    \small
    \scriptsize
    \resizebox{.39\textwidth}{!}{%
    \begin{tabular}{|c|cc|c|c|c|}
    \hline
         \multicolumn{1}{|c}{} & \multicolumn{2}{|c|}{threshold} & economic & \multicolumn{2}{c|}{gender} \\ 
         \multicolumn{1}{|c}{} & \multicolumn{2}{|c|}{maximizing}  & benefit (RMB) & \multicolumn{1}{c}{TF} & \multicolumn{1}{c|}{OF} \\
         \hline
DNN  & A: & (default) & 2.91$\pm$0.14  & $92.0$ & $55.2$ \\ 
             & B: & econ & \textbf{3.17$\pm$0.31}  & $33.7$ & $70.7$ \\ 
             & C: & TF & 2.93$\pm$0.09  & $\mathbf{98.1}$ & $58.8$ \\ 
             & D: & OF & 2.95$\pm$0.15  & $54.6$ & $\mathbf{76.9}$ \\ 
         \hline
    \end{tabular}} \\[10pt]
    \label{app_tab:blood_mult_thresh_dnn}
    \end{centering}
\end{table}

Figure \ref{app_fig:blood_gend_2d_fairness_stars} depicts where many of these different thresholds end up on a 2D chart plotting $TF$ v. $OF$.
Each dot corresponds to a program evaluation of a specific threshold, and interestingly the overall shape seems to follow a triangular curve.
More blue refers to a greater percentage of men being treated; more red refers to a greater percentage of women being treated.
On the right side of the figure, we see the purple corner corresponding to $100\%$ treatment fairness --an equal proportion of men and women receiving the promotion.
In order to achieve outcome fairness, however, we need to keep treating a larger percentage of the male population.
This is due to the fact that in this dataset, men were more likely to donate blood when offered a grocery coupon.
Consequently, to balance this, outcome fairness says we should treat less women and more men to counteract the fact that men are more likely to donate than women.
Balancing the opportunity of a coupon with the burden of donating blood and debating which of these is the `correct' decision is at the heart of what the fairness literature has been disputing over.
Directly visualizing this tradeoff in a practical setting is an exciting result of our setup.

We can attempt to utilize different thresholds to optimize for each of the three different objectives.
Looking back at Figure \ref{app_fig:blood_gend_2d_fairness_stars} we see markers labeled A, B, C, and D corresponding to the thresholds used in Table \ref{app_tab:blood_mult_thresh_dnn} where we can see how each of these thresholds performed on a held-out testing set.
Indeed, our random experiments auditing set is able to provide a sufficiently accurate estimate of how each different threshold will influence the outcome in the real-world.

\end{document}